\documentclass[letterpaper, 10pt, conference]{ieeeconf}  

\IEEEoverridecommandlockouts                              

\usepackage{xr}
\newcommand{\omitted}[1]{}
\usepackage{subfigure}
\usepackage{float}
\usepackage{graphicx}    
\usepackage{xspace}
\usepackage[font=small,labelfont=bf]{caption}
\usepackage{siunitx}
\usepackage[absolute]{textpos}

\newcommand{\extended}[2]{{#2}\xspace} %

\usepackage{comment}
\usepackage{siunitx}
\usepackage{relsize}
\usepackage{ifthen}
\usepackage[colorinlistoftodos]{todonotes}

\usepackage[caption=false]{subfig}

\usepackage[vlined,ruled,linesnumbered]{algorithm2e}
\usepackage{graphics} 
\usepackage{rotating}
\usepackage{color}
\usepackage{enumerate}
\usepackage[T1]{fontenc}
\usepackage{psfrag}
\usepackage{epsfig} 
\usepackage{booktabs}
\usepackage{graphicx,url}
\usepackage{multirow}
\usepackage{array}
\usepackage{latexsym}
\usepackage{amsfonts}
\usepackage{amsmath}
\usepackage{amssymb}
\usepackage{xstring}
\usepackage[noend]{algorithmic}
\usepackage{multirow}
\usepackage{xcolor}
\usepackage{prettyref}
\usepackage{flexisym}
\usepackage{bigdelim}
\usepackage{breqn} 
\usepackage{listings}
\let\labelindent\relax
\usepackage{enumitem}
\usepackage{xspace}
\usepackage{bm}
\graphicspath{{./figures/}}
\usepackage{tikz}
\usetikzlibrary{matrix,calc}

\usepackage{mdwlist}

\let\itemize\undefined
\makecompactlist{itemize}{stditemize}

\newrefformat{prob}{Problem\,\ref{#1}}
\newrefformat{def}{Definition\,\ref{#1}}
\newrefformat{sec}{Section\,\ref{#1}}
\newrefformat{sub}{Section\,\ref{#1}}
\newrefformat{prop}{Proposition\,\ref{#1}}
\newrefformat{app}{Appendix\,\ref{#1}}
\newrefformat{alg}{Algorithm\,\ref{#1}}
\newrefformat{cor}{Corollary\,\ref{#1}}
\newrefformat{thm}{Theorem\,\ref{#1}}
\newrefformat{lem}{Lemma\,\ref{#1}}
\newrefformat{fig}{Fig.\,\ref{#1}}
\newrefformat{tab}{Table\,\ref{#1}}

\newtheorem{theorem}{Theorem}

\newtheorem{assumption}[theorem]{Assumption}

\newcommand{\bdmath}{\begin{dmath}}
\newcommand{\edmath}{\end{dmath}}
\newcommand{\beq}{\begin{equation}}
\newcommand{\eeq}{\end{equation}}
\newcommand{\bdm}{\begin{displaymath}}
\newcommand{\edm}{\end{displaymath}}
\newcommand{\bea}{\begin{eqnarray}}
\newcommand{\eea}{\end{eqnarray}}
\newcommand{\beal}{\beq \begin{array}{ll}}
\newcommand{\eeal}{\end{array} \eeq}
\newcommand{\beas}{\begin{eqnarray*}}
\newcommand{\eeas}{\end{eqnarray*}}
\newcommand{\ba}{\begin{array}}
\newcommand{\ea}{\end{array}}
\newcommand{\bit}{\begin{itemize}}
\newcommand{\eit}{\end{itemize}}
\newcommand{\ben}{\begin{enumerate}}
\newcommand{\een}{\end{enumerate}}

\newcommand{\calC}{{\cal C}}
\newcommand{\calD}{{\cal D}}
\newcommand{\calE}{{\cal E}}

\newcommand{\calJ}{{\cal J}}

\newcommand{\setal}{~\emph{et~al.}\xspace}
\newcommand{\eg}{\emph{e.g.,}\xspace}
\newcommand{\ie}{\emph{i.e.,}\xspace}
\newcommand{\myParagraph}[1]{{\bf #1.}\xspace}

\newcommand{\M}[1]{{\bm #1}} 
\renewcommand{\boldsymbol}[1]{{\bm #1}}

\newcommand{\hide}[1]{}

\newcommand{\hiddenText}{{\color{gray} hidden text.}}
\newcommand{\hideWithText}[1]{\hiddenText}

\newcommand{\subject}{\text{ subject to }}

\DeclareMathOperator*{\argmin}{arg\,min}

\newcommand{\tran}{^{\mathsf{T}}}

\newcommand{\diag}[1]{\mathrm{diag}\left(#1\right)}
\newcommand{\trace}[1]{\mathrm{tr}\left(#1\right)}

\newcommand{\e}{{\mathrm e}}
\newcommand{\inv}{^{-1}}

\newcommand{\zero}{{\mathbf 0}}
\newcommand{\eye}{{\mathbf I}}

\newcommand{\Real}[1]{ { {\mathbb R}^{#1} } }

\newcommand{\at}[1]{^{(#1)}}

\newcommand{\SOtwo}{\ensuremath{\mathrm{SO}(2)}\xspace}
\newcommand{\SOthree}{\ensuremath{\mathrm{SO}(3)}\xspace}

\newcommand{\MJ}{\M{J}}

\newcommand{\MG}{\M{G}}

\newcommand{\MR}{\M{R}}

\newcommand{\MX}{\M{X}}
\newcommand{\MY}{\M{Y}}

\newcommand{\vb}{\boldsymbol{b}}

\newcommand{\ve}{\boldsymbol{e}}
\newcommand{\vf}{\boldsymbol{f}}
\newcommand{\vg}{\boldsymbol{g}}

\newcommand{\vl}{\boldsymbol{l}}

\newcommand{\vo}{\boldsymbol{o}}
\newcommand{\vp}{\boldsymbol{p}}
\newcommand{\vq}{\boldsymbol{q}}

\newcommand{\vxx}{\boldsymbol{x}} 
\newcommand{\vy}{\boldsymbol{y}}
\newcommand{\vw}{\boldsymbol{w}}

\newcommand{\vtau}{\boldsymbol{\tau}}

\newcommand{\blue}[1]{{\color{blue}#1}}

\newcommand{\linkToPdf}[1]{\href{#1}{\blue{(pdf)}}}
\newcommand{\linkToPpt}[1]{\href{#1}{\blue{(ppt)}}}
\newcommand{\linkToCode}[1]{\href{#1}{\blue{(code)}}}
\newcommand{\linkToWeb}[1]{\href{#1}{\blue{(web)}}}
\newcommand{\linkToVideo}[1]{\href{#1}{\blue{(video)}}}
\newcommand{\linkToMedia}[1]{\href{#1}{\blue{(media)}}}
\newcommand{\award}[1]{\xspace} 

\graphicspath{{./figures/}}

\DeclareMathOperator{\sign}{sgn}

\newcommand{\finalTime}{{t_f}}
\newcommand{\graspTime}{{t_g}}

\newcommand{\quadpos}{\vp}
\newcommand{\quadrot}{\MR}
\newcommand{\quadrvel}{\boldsymbol{\Omega}}
\newcommand{\propforces}{\vf}
\newcommand{\load}{load}
\newcommand{\COM}{CM}
\newcommand{\lenCOM}{L_{\COM}}
\newcommand{\nrNodes}{N}
\newcommand{\nodes}{\MY}
\newcommand{\node}{\vy}
\newcommand{\MXbar}{\bar{\MX}}
\newcommand{\MYbar}{\bar{\MY}}

\newcommand{\Force}{\mathbf{F}}
\newcommand{\restdisp}{\bar{\Delta}\nodes_{ijkl}}
\newcommand{\defdisp}{\Delta\nodes_{ijkl}}
\newcommand{\defgrad}{\MG}
\newcommand{\restvol}{\bar{v}}
\newcommand{\voldef}{v_F}
\newcommand{\nodemass}{{m}}

\newcommand{\piola}{\mathbf{P_s}}
\newcommand{\meshforce}{\Force_{mesh}}
\newcommand{\idx}{r}

\newcommand{\routing}{\nodes^{t_i}}
\newcommand{\restlengths}{\vl}
\newcommand{\restlength}{l}
\newcommand{\tendonforce}{\Force_{tendon}}
\newcommand{\tension}{\sigma}
\newcommand{\tensions}{\boldsymbol{\sigma}}
\newcommand{\tendondef}{\gamma}

\newcommand{\pinforce}{\Force_{pin}}
\newcommand{\gravityforce}{\Force_{gravity}}

\newcommand{\dt}{\text{d}t}
\newcommand{\softCost}{\calC}
\newcommand{\rotcol}{\vb}
\newcommand{\rotcolx}{\vb_x}
\newcommand{\rotcoly}{\vb_y}
\newcommand{\rotcolz}{\vb_z}

\newcommand{\loadml}{m_L}
\newcommand{\fdes}{\mathbf{f_d}}

\newcommand{\youngModulus}{\calE}
\newcommand{\poissonRatio}{\nu}

\newcommand{\Properties}{Energy}

\PassOptionsToPackage{end}{algorithmic}

\renewcommand{\linkToPdf}[1]{\xspace}
\renewcommand{\linkToPpt}[1]{\xspace}
\renewcommand{\linkToCode}[1]{\xspace}
\renewcommand{\linkToWeb}[1]{\xspace}
\renewcommand{\linkToVideo}[1]{\xspace}
\renewcommand{\linkToMedia}[1]{\xspace}
\renewcommand{\award}[1]{\xspace}
\newcommand{\myhspace}{\hspace{0mm}}
\newcommand{\mpw}{4.2cm}

\title{\LARGE \bf Control and Trajectory Optimization for Soft Aerial Manipulation}

\author{Joshua Fishman and Luca Carlone
\thanks{J.\,Fishman is with the Dept. of Mechanical Engineering, Massachusetts Institute of Technology,
        Cambridge, MA, {\tt\footnotesize joshuaf@mit.edu}}%
\thanks{L.\,Carlone is with the Dept. of Aeronautics and Astronautics, Massachusetts Institute of Technology,
        Cambridge, MA, {\tt\footnotesize lcarlone@mit.edu}}%
}

\begin{document}

\maketitle

\begin{textblock}{10}(3,0.1)
\large \centering
\noindent Please cite as follows: J. Fishman and L. Carlone, 

\noindent "Control and Trajectory Optimization for Soft Aerial Manipulation"

\noindent IEEE Aerospace Conference, 2021
\end{textblock}

\thispagestyle{empty}
\pagestyle{empty}

\renewcommand{\baselinestretch}{0.90}

\begin{abstract}


Manipulation and grasping with unmanned aerial vehicles (UAVs) currently require accurate positioning and are often executed at reduced speed to ensure successful grasps.
This is due to the fact that typical UAVs can only accommodate  
rigid manipulators with few degrees of freedom, which limits their capability to compensate for disturbances 
caused by the vehicle positioning errors. Moreover, {UAVs} have to minimize external contact forces 
in order to maintain stability. 
Biological systems, on the other hand, exploit softness to overcome similar limitations, and 
{leverage} compliance to enable aggressive grasping.
This paper investigates control and trajectory optimization for a \emph{soft aerial manipulator}, consisting of a quadrotor and a tendon-actuated soft gripper, in which the advantages of softness can be fully exploited.
To the best of our knowledge, this is the first work at the intersection between soft manipulation 
\mbox{and UAV control.
We present} a decoupled approach for the quadrotor and the soft gripper, combining 
(i) a  geometric controller and a minimum-snap trajectory optimization for the quadrotor (rigid) base, with 
(ii) a quasi-static finite element model and control-space interpolation for the soft gripper. 
We prove that the geometric controller asymptotically stabilizes the quadrotor velocity and attitude
despite the addition of the soft load. Finally, we evaluate the proposed system in a realistic soft dynamics simulator, and show that:
(i) the geometric controller is fairly insensitive to the soft payload, 
(ii) the platform can reliably grasp unknown objects despite inaccurate positioning and initial conditions, and
(iii) the decoupled controller is amenable for real-time execution.
\end{abstract}

Video Attachment: 
\blue{\footnotesize\url{https://youtu.be/NNpQxP0SPFk}}

\section{Introduction}
\label{sec:intro}

\extended{Aerial manipulation is a fundamental capability for autonomous systems and has the potential to unleash several applications, ranging from autonomous transportation and construction, 
delivery of medical goods,
precision agriculture,
and infrastructure monitoring,
among others~\cite{Khamseh18ras-aerialManipulationSurvey}.}
{
Aerial manipulation is a fundamental capability for autonomous systems and has the potential to unleash several applications, including autonomous transportation and construction~\cite{Loianno18ral}, 
 medical goods delivery~\cite{Thiels15amj},
\extended{agriculture and forestry~\cite{Ore13fsr},}
{agriculture and forestry (water sampling~\cite{Ore13fsr}, forest canopy sampling~\cite{Kaslin18front}),}
infrastructure monitoring and maintenance~\cite{Bodie19arxiv},
\extended{and autonomous charging~\cite{Moore09iaa},}{and autonomous charging via perching~\cite{Moore09iaa},} 
among others~\cite{Khamseh18ras-aerialManipulationSurvey}.
}

Quadrotors have been extensively used and investigated as platforms for navigation and inspection~\cite{Falanga18ral,Bodie19arxiv}, due to their  versatility and maneuverability. However, they impose several constraints when it comes to manipulation. 
First, small quadrotors (often called \emph{micro aerial vehicles}~\cite{Loianno18ral}) have limited payload, hence they can only carry relatively simple and lightweight manipulators.
This intrinsically limits their capability to compensate for disturbances,
such as the ones caused by the vehicle positioning errors during grasp execution. 
Second, aerial systems are inherently fragile and imprecise~\cite{Dollar10ijrr}; 
for this reason, external contact forces arising from unplanned contacts have to be avoided in order to preserve stability~\cite{Khamseh18ras-aerialManipulationSurvey}.
Many works circumvent these issues by reducing the speed and accelerations of the quadrotor~\cite{Dentler16cca-aerialManipulation,Rossi17ral}. 
This reduces the magnitude of external contact forces but has the drawback of making operation inefficient, especially considering the short flight time of small quadrotors. 
The work~\cite{Thomas14bioinspiration} demonstrates agile grasping but the problem setup is simplified to avoid unplanned contact forces (\ie the object to grasp is suspended rather than lying on a surface).  
The recent literature bears witness to an increasing interest in compliant aerial manipulation.
Compliance and under-actuation are now widely exploited, 
to enable the manipulation of objects of varying shape and to minimize disturbances imposed by the environment~\cite{Backus14iros}. However, to our knowledge this has been restricted to cases where the compliant elements either have limited degrees of freedom (often due to payload constraints) or do not affect the quadrotor dynamics (\eg cable-slung loads).

\begin{figure}[t]
\vspace{-5mm}
    \centering
    \includegraphics[width=3.5in]{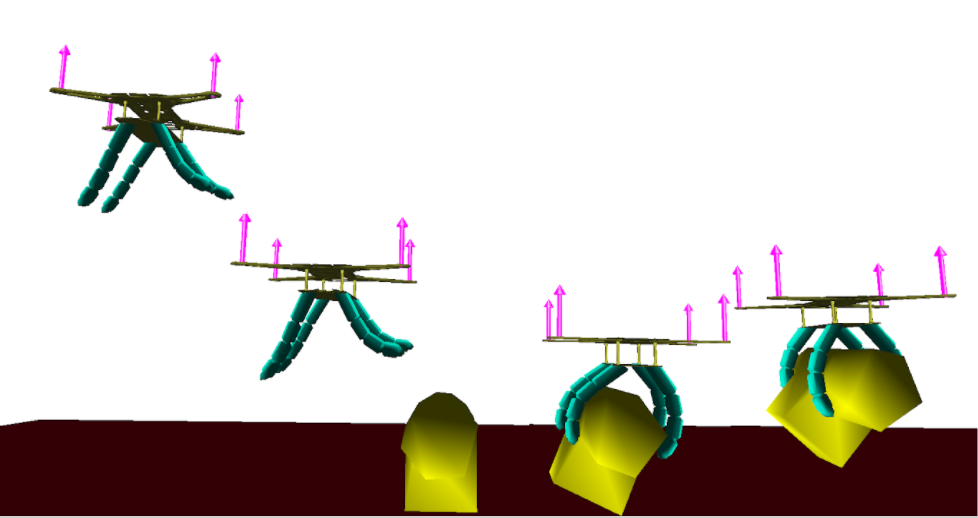}
    \caption{We investigate control and trajectory optimization for a \emph{soft aerial manipulator}, consisting of a quadrotor 
    (yellow frame with propeller thrusts in magenta) and a tendon-actuated soft gripper (cyan). The figure shows a temporal sequence leading to a successful grasp in a realistic soft dynamics simulator.
    \label{fig:timelapse}\vspace{-5mm}}  
\end{figure}

On the other hand, soft materials are ubiquitous in nature and enable the performance and robustness which so differentiate natural from artificial systems~\cite{Bern19rss}. 
Soft manipulators passively conform to the grasped object, enabling tolerance to imprecisions  and reducing the need for explicit grasp analysis; this is an example of \emph{morphological computation}, the exploitation of passive mechanical elements to supplement explicit control~\cite{Rus15nature}. 
Moreover, a soft gripper can be realized using lightweight materials (\eg foam~\cite{Schlagenhauf18humanoids}), making it a viable option for small UAVs.
Despite the potential to use soft grippers as a lightweight and compliant alternative for aerial manipulation, little attempt has yet been made to explicitly model and control continuously deformable, soft structures in an aerial context. In general, such soft elements are continuously deformable and possess theoretically infinite degrees of freedom.  
Therefore, they cannot be modeled in closed form and are not differentially flat, putting them at odds with typical techniques for UAV control. 


This paper bridges the quadrotor control and planning literature with the growing field of soft robotics.
In particular,  we present a soft aerial manipulator (Fig.~\ref{fig:timelapse}) and 
investigate control and trajectory optimization algorithms to enable aggressive grasping of an object lying on a surface.

\begin{figure}[t]
    \begin{center}
    \begin{minipage}{\textwidth}
    \begin{tabular}{cc}%
    \myhspace \hspace{-3mm}
            \begin{minipage}{\mpw}%
            \centering%
            \includegraphics[trim=0mm 0mm 0mm 5mm, clip,height=2.7cm]{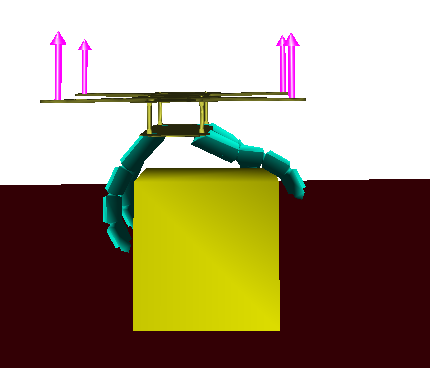} \\
            \hspace{0cm}(a) 
            \end{minipage}
        & \myhspace 
            \begin{minipage}{\mpw}%
            \centering%
            \includegraphics[trim=0mm 0mm 0mm 5mm, clip,height=2.7cm]{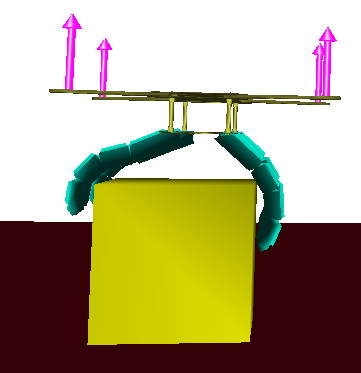} \\
            \hspace{0cm}(b) 
            \end{minipage}
        \end{tabular}
    \end{minipage}
    \vspace{-2mm} 
    \caption{Morphological computation refers to the ability of a physical system to supplement explicit control. Here, we show our soft gripper achieving successful grasps despite significant variations in the quadrotor position; the gripper command is the same in (a)-(b).\label{fig:adaptive_grasp}\vspace{-7mm} }
    \end{center}
\end{figure}

After reviewing the literature in Section~\ref{sec:relatedWork}, 
Section~\ref{sec:overview} describes the proposed soft aerial manipulator and states the grasping problem in terms of the concurrent planning and control of both the rigid quadrotor base  and the soft gripper. 
The section also provides an overview of the proposed algorithmic approach, based on decoupling the 
control and planning for the rigid and soft components; this is made possible by the insight in Theorem~\ref{thm:controllerConvergence} and the resilience to positioning errors afforded by the soft gripper (Fig.~\ref{fig:adaptive_grasp}). 

Section~\ref{sec:softGripper} describes the control law and the trajectory optimization approach for the soft gripper.
We assume that the gripper remains in quasi-static equilibrium and compute forward kinematics by minimizing total energy (defined via finite element methods) using Newton's method.  \extended{We then compute optimal tendon control with a gradient descent methodology,  assuming that the quadrotor attains its nominal trajectory and relying on the inherent adaptability of the soft gripper to compensate for deviations (again, Fig.~\ref{fig:adaptive_grasp}).}{We then compute optimal tendon control with a gradient descent methodology and linearly interpolate these over the length of the trajectory. In defining an objective function for the gripper we assume that the quadrotor attains its nominal trajectory, relying on the inherent adaptability of the soft gripper to compensate for deviations (again, Fig.~\ref{fig:adaptive_grasp}).}

Section~\ref{sec:quadrotorBase} reviews a standard geometric controller for a quadrotor. The novel insight here is that by modeling the gripper as a symmetric soft payload and treating torque imposed by it as a disturbance, we can prove that the geometric controller stabilizes the quadrotor velocity and attitude irrespective of the soft gripper (Theorem~\ref{thm:controllerConvergence}). Moreover, we use minimum-snap trajectory optimization and add an intermediate waypoint to support the grasp, while abstracting away the manipulation aspects 
(accounted for by the gripper).

Section~\ref{sec:experiments} presents numerical experiments performed in a realistic soft dynamics simulator, \emph{SOFA}~\cite{Faure12sofaam}.
The experiments highlight the effectiveness of the proposed system and show that:
(i) the geometric controller is fairly insensitive to the soft payload, 
(ii) the platform can reliably grasp unknown objects despite inaccurate positioning and starting from a variety of initial conditions, and
(iii) the decoupled controller is amenable for real-time execution.

\section{Related Work}
\label{sec:relatedWork}

\myParagraph{Aerial Manipulation}
Aerial manipulation with under-actuated and un-actuated components has been subject of extensive research; however, this has been restricted to components with limited degrees of freedom and in which the combined system is differentially flat.
Cable-slung loads (F{\"o}hn\setal~\cite{Foehn17rss}, Sreenath\setal~\cite{Sreenath13icra}) are a well-studied under-actuated payload, but are differentially flat and 
are typically designed not to impose a torque on the aerial platform. The continuously deformable cable is either treated as massless or reduced to a finite number of links (Goodarzi\setal~\cite{Goodarzi14acc}). 
Thomas\setal~\cite{Thomas14bioinspiration} carry out aggressive aerial manipulation inspired by birds of prey using an under-actuated gripper, which can potentially grasp complex objects 
(see also Pounds\setal~\cite{Pounds11icra} and Backus\setal~\cite{Backus14iros}). However, 
the work~\cite{Thomas14bioinspiration} focuses on the case of a suspended object, 
which avoids unplanned collisions
 through the gripper. 
The AEROARMS project (Caballero\setal~\cite{Cabellero18iros}) explored the use of a manipulator with a flexible link to minimize disturbance on the aerial platform, but did not model or account for its impact on the drone dynamics.
Yuksel\setal~\cite{yuksel16iros} show differential flatness for aerial manipulators with arbitrary but finite numbers of  rigid or compliant joints; this result does not hold for the continuously deformable case. 

\myParagraph{Morphing Drones}
While not being targeted at manipulation, a recent set of papers investigates  the design of UAV platforms that can mechanically change shape to tolerate collisions or fit into narrow gaps.
{Mintchev\setal~\cite{Mintchev17ral} use insect-inspired structural compliance in a quadrotor frame to minimize impact damage, but this compliance does not affect the quadrotor dynamics during flight.} 
Falanga\setal~\cite{Falanga18ral} develop a drone that can fold its arms to fit into narrow gaps;
morphological adaptability is limited to the plane so that the resulting dynamics can be expressed in closed form. 
Ramon\setal~\cite{Ramon19iros} propose soft landing gear (similar in spirit to our design), but do not model the soft component nor its interaction with the quadrotor controller or control the soft landing gear beyond a binary open/close; moreover, the work focuses on landing
 rather than manipulation. Deng\setal~\cite{Deng20arxiv} (published after our preprint \cite{Fishman20arxiv-softDrone}) implement and control a soft-bodied multicopter in simulation; their work relies on a gray-box neural network for system identification and does not consider manipulation.
Other related work includes quadrotors with tilting body or propellers 
(Ryll\setal~\cite{Ryll13icra-tiltingPropellers,Ryll15tro-tiltingPropellers}, Hintz\setal~\cite{Hintz14icuas-morphingDrone}, Riviere\setal~\cite{Riviere18softRobotics-morphingDrone}), 
scissor-like foldable quadrotors (Zhao\setal~\cite{Zhao17iros-deformableDrones}), 
 quadrotors with sprung-hinge-based foldable arms (Bucki and Mueller~\cite{Bucki19icra-morphingDrones}),
 and small-winged drones with morphing wing design (Di Luca\setal~\cite{DiLuca17interfaceFocus-morphingWings}). 
Contrarily to these works we consider a quadrotor carrying a soft gripper and prove that such a payload does not destroy the asymptotic stability of an existing geometric controller by 
Lee\setal~\cite{Lee10cdc-geometricControl}.

\myParagraph{Soft Robotics}
Continuously deformable, entirely compliant robots represent the extremum of the trend towards compliance and under-actuation.  However, traditional rigid-body modeling and control techniques fall short when confronted with infinite degrees of freedom. The emerging discipline of soft robotics has developed principled approaches to allow control of these systems, opening a new frontier in manipulation. 
Rus and Tolley~\cite{Rus15nature} and Thuruthel\setal~\cite{Thuruthel18softrobotics} provide a comprehensive review 
of soft robotics and soft manipulation.
King\setal~\cite{King18humanoids}, Manti\setal~\cite{Manti15SoftRobotics}, and Hassan\setal~\cite{Hassan15embc} design bio-inspired tendon-actuated soft grippers. Marchese\setal~\cite{Marchese14icra,Marchese15icra} implement kinematics and trajectory optimization for hydraulic soft manipulators based on piecewise-constant-curvature approximations. Bern\setal~\cite{Bern17iros,Bern19rss} and Duriez\setal~\cite{Duriez13icra} model the kinematics and dynamics of tendon-actuated soft robots using finite element methods. 
These works have not been applied in the context of aerial manipulation, where the soft manipulator becomes a time-varying payload for the UAV and impacts its dynamics.

\section{Soft Aerial Manipulation: Problem Statement and Decoupled Approach}
\label{sec:overview}

\subsection{System Overview and Problem Statement}
\label{sec:problemFormulation}  

\begin{figure}
    \begin{minipage}{\textwidth}
    \begin{tabular}{cc}%
    \myhspace \hspace{-3mm}
            \begin{minipage}{1.5cm}%
            \centering \vspace{-0.8cm}
            (a) Top view \vspace{1.2cm}\\
            (b) Side view
            \end{minipage}
        &  \hspace{-3mm}
            \begin{minipage}{7cm}%
            \centering%
            \includegraphics[width=0.8\textwidth]{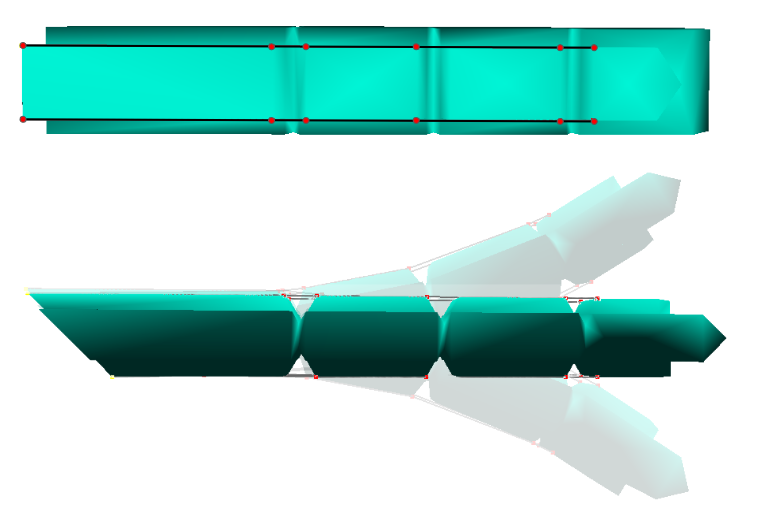} 
            \end{minipage}
        \end{tabular}
        \vspace{-3mm}
    \end{minipage}
    \caption{Soft finger with two pairs of tendons on opposite sides. Tendons (black) pass through a set of nodes (red) attached to the finger, such that pulling a tendon causes a contraction of the finger.\label{fig:finger} \vspace{-7mm}}
\end{figure}

Our soft aerial manipulator (Fig.~\ref{fig:adaptive_grasp}) comprises the frame of a standard quadrotor with the (rigid and heavy) landing gear replaced by a soft gripper.
The gripper consists of four soft fingers and is based on the design by Hassan\setal~\cite{Hassan15embc}. Each finger is attached to the quadrotor base and actuated with two pairs of tendons on opposite sides (Fig.~\ref{fig:finger}).

Each finger is actuated by setting a desired length  (later called the \emph{rest length}) at the tendons. Similarly to \cite{Bern19rss}, actuation of a pair of tendons (lying on the same side of the finger) is coupled to prevent finger twist.
The quadrotor base uses four motors and propellers for actuation, as usual.
In summary, the system uses 12 control variables (four motor speeds for the quadrotor, and one for each pair of tendons on the two sides of the fingers) to control an infinite-dimensional state (including the finite-dimensional quadrotor state 
and the infinite-dimensional state describing the configuration of the soft gripper). 
The soft gripper model is described in Section~\ref{sec:softGripper}, while 
the quadrotor model is given in Section~\ref{sec:quadrotorBase}.

Our manipulator is tasked with grasping an object of unknown shape lying over an unknown surface: 
the system is only provided with the centroid of the object. 
In particular, we are interested in (i) computing a trajectory of state variables over time, and (ii) developing a control law that can track the computed trajectory to ensure
a successful grasp in the face of external disturbances. 
We assume we can measure the full state of the quadrotor (its 3D pose and linear and angular velocities, typically observable using a motion capture system~\cite{Mellinger11icra} or visual-inertial state estimation~\cite{Loianno16icra-vioQuadrotors}), while we operate the soft gripper in open loop (\ie our approach does not need to measure the state of the gripper). 
While our goal is to simultaneously obtain trajectories and controllers for the soft gripper and the rigid quadrotor,
in the following we propose a decoupled approach that implements separate planners/controllers for both subsystems. 

\subsection{Decoupled Control and Trajectory Optimization}
\label{sec:decoupledApproachOverview}

Let us call $\MX(t)$ the state of the quadrotor base (\ie a 3D pose and its derivatives) at time $t$, 
and $\MY(t)$ the infinite-dimensional matrix describing the 3D position of every point of the soft gripper.
Moreover, call 
$\propforces(t)$ 
the quadrotor propeller {thrust} forces at time $t$, and 
$\restlengths(t)$
the tendon rest lengths that actuate the fingers. 
To simplify the notation, 
below we omit the dependence on time $t$ when possible. 

The \emph{soft aerial grasping problem} considered in this paper can be formulated as an optimal control problem:
\begin{equation}
\begin{array}{rcl}
\hspace{-2mm} 
    (\MX^\star, \MY^\star, \propforces^\star, \restlengths^\star)  \label{eq:optControl}
     = \hspace{-9mm}& \hspace{-2mm}
     \displaystyle \argmin_{\MX, \MY, \propforces, \restlengths} &
    \int_{0}^{\finalTime}  \calJ(\MX, \MY, \propforces, \restlengths) \dt
     \\
    & \hspace{-4mm}  \subject & 
        \calD(\MX, \MY, \propforces, \restlengths) = 0 \\
    && \MX(0) = \MXbar_0, \quad \MY(0) = \MYbar_0 \\ 
    && \MX(\finalTime) = \MXbar_\finalTime, \; \MY(\finalTime) = \MYbar_\finalTime \\
     && \MY(\graspTime) = \MYbar_\graspTime
\end{array}
\end{equation}
where $\calJ(\MX, \MY, \propforces, \restlengths)$ is the cost functional that, for instance, penalizes 
control usage or encourages smooth state changes,
 the constraint $\calD(\MX, \MY, \propforces, \restlengths)=0$ ensures that the solution satisfies 
 the platform dynamics,
$(\MXbar_0,\MYbar_0)$ is the given initial state of the soft aerial manipulator at the initial time $t=0$, 
$(\MXbar_\finalTime,\MYbar_\finalTime)$ is the desired state at the final time $t_f$ (say, the end of the execution), 
and $\MYbar_\graspTime$ is the desired state of the soft gripper at the time of grasp $\graspTime \in [0,\finalTime]$.
In words, Problem~\eqref{eq:optControl} looks for minimum-cost controls such that 
 the platform moves from an initial to a final state, 
and the soft gripper is in a suitable configuration during grasp.

While in principle one would like to obtain a \emph{control policy} that computes a suitable control
$(\propforces, \restlengths)$ for every possible state, doing so is hard even without a soft gripper.  
Therefore, related work solves problems akin to~\eqref{eq:optControl} by first performing 
\emph{trajectory optimization}, \ie 
computing an open loop state trajectory and then designing a controller that tracks such a trajectory~\cite{Lee10cdc-geometricControl}. 
We follow the same approach and decouple the optimal control problem into trajectory optimization and tracking control. 
However, we are still left with the complexity that our soft aerial manipulator is not differentially 
flat, which is a key requirement for tractable trajectory optimization in related work~\cite{Mellinger11icra}.
To circumvent this issue, we further decouple trajectory optimization and control for the 
 quadrotor base and the soft gripper as follows.

We split Problem~\eqref{eq:optControl} into the cascade of two problems. 
First,  we solve the {\bf drone control subproblem}, 
where we look for an optimal control action for the drone propeller forces $\propforces$ 
while
treating the soft payload as 
an unknown disturbance. This can be formulated as follows:  
\begin{equation}
\label{eq:optControlQuadrotor}
\begin{array}{rcl}
    (\MX^\star, \propforces^\star)  = &  \argmin_{\MX,\propforces} & \int_{0}^{\finalTime}  \calJ_q(\MX, \propforces) \dt \\
    & \subject &   \calD_q(\MX, \MY, \propforces)=0 \\
    && \MX(0) = \MXbar_0, \quad \MX(\finalTime) = \MXbar_\finalTime, \\ 
    && \MX(\graspTime) = \MXbar_\graspTime
\end{array}
\end{equation}
where $\calJ_q$ and $\calD_q$ now only involve the quadrotor state and dynamics, and where we 
relaxed the 
grasp condition $\MY(\graspTime) = \MYbar_\graspTime$ in~\eqref{eq:optControl}, with a condition on the state of the quadrotor during the grasp $\MX(\graspTime) = \MXbar_\graspTime$. 
Intuitively, the drone has to ensure it is close enough to the object at time $t_g$ to enable the soft gripper to grasp, but without worrying about the specific configuration of the gripper. 
Note how the drone dynamics are a function of the soft gripper configuration $\MY$, which is treated as 
an unknown disturbance, hence will not used to solve~\eqref{eq:optControlQuadrotor} (see Section~\ref{sec:quadrotorBase}).

After solving~\eqref{eq:optControlQuadrotor} and obtaining the nominal (open loop) quadrotor trajectory $\MX^\star$, 
we solve the {\bf soft-gripper control subproblem}, 
where we look for an optimal control action for the tendons rest lengths $\restlengths$: 
\begin{equation}
\label{eq:optControlSoftGripper}
\begin{array}{rcl}
    (\MY^\star, \restlengths^\star)
     = &   \argmin_{\restlengths} &
    \int_{0}^{\finalTime}  \calJ_s(\MX^\star, \MY, \restlengths) \dt
     \\
    & \hspace{-4mm}  \subject & 
       \calD_s(\MX^\star, \MY, \restlengths) = 0 \\
    && \MY(0) = \MYbar_0, \quad \MY(\finalTime) = \MYbar_\finalTime(\MX^\star),\\
     && \MY(\graspTime) = \MYbar_\graspTime(\MX^\star)
\end{array}
\end{equation}
where now the soft-gripper dynamics $\calD_s(\MX^\star, \MY, \restlengths)$, the grasp configuration $\MYbar_\graspTime(\MX^\star)$,
and the terminal state $\MYbar_\finalTime(\MX^\star)$ depend on the (fixed) nominal drone trajectory $\MX^\star$. 
In practice the object to be grasped has unknown shape and the soft gripper has an infinite number of points, hence it is unrealistic to enforce the condition $\MY(\graspTime) = \MYbar_\graspTime(\MX^\star)$; 
{in Section~\ref{sec:graspObjectives} we will replace such a condition with a more realistic one 
involving only the
positions of the 4 fingertips and the object centroid.}

In the following sections, we describe our choice of cost functions and 
discuss how to attack problems~\eqref{eq:optControlQuadrotor} and \eqref{eq:optControlSoftGripper}, using tools from quadrotor control~\cite{Lee10cdc-geometricControl} and soft robotics~\cite{Bern17iros}.   
\section{Open-loop Control and Trajectory Optimization for a Soft Gripper}
\label{sec:softGripper}

This section describes how to solve the soft-gripper subproblem~\eqref{eq:optControlSoftGripper} for a specific choice of cost function. {We make the following key assumption.
\begin{assumption}[Quasi-static approximation]
\label{ass:softGripper}
The soft gripper is \emph{quasi-static}, \ie there is an instantaneous relation between rest lengths and gripper configuration.
\end{assumption}

This allows simplifying the model by neglecting the soft gripper dynamics and is fairly common in soft robotics~\cite{Bern17iros}.}


\subsection{Objectives for Aggressive Soft Grasping}
\label{sec:graspObjectives}

This section discusses our choice of objective function in eq.~\eqref{eq:optControlSoftGripper}.
As anticipated in Section~\ref{sec:overview}, rather than matching a desired configuration $\MYbar_\graspTime$ (which would be hard to compute without knowing the shape of the object to grasp), we prefer  finding a control which optimizes the 
positions of the fingertips with respect to the centroid of the object we want to grasp.
Mathematically, we drop the constraint $\MY(\graspTime) = \MYbar_\graspTime(\MX^\star)$ in eq.~\eqref{eq:optControlSoftGripper}
and optimize an objective which is only 
function of the 
fingertip positions $\node_{tip_i}$ (each being a  point in $\MY$) and the centroid $\vo$ of the object we want to grasp.
This is an exemplary instantiation of the idea of \emph{morphological computation}: rather than planning the full configuration of the soft gripper, we only plan for the fingertips and have the softness of the fingers adjust to the shape of the object we want to grasp.

Specifically, we plan grasps in two phases, ``approach'' and ``grasp''. Intuitively, the ``approach'' phase involves opening the gripper as the quadrotor approaches the target location so as to allow the fingertips to surround the grasped object, while the ``grasp'' phase involves closing the gripper, contracting the fingertips to achieve an enveloping grasp. 
Mathematically, this reduces to assuming the following form for the objective function in eq.~\eqref{eq:optControlSoftGripper}:
\bea
\label{eq:softObjective}
\!\!\!
\int_{0}^{\finalTime}  \calJ_s(\MX^\star, \MY, \restlengths) \dt &=& 
\int_{0}^{\finalTime}  \calJ_{\text{tendon}}(\MX^\star, \MY, \restlengths) \dt \nonumber  \\
     &+&  \softCost_{\text{approach}} (\nodes(t_a))\nonumber  \\
     &+&  \softCost_{\text{grasp}} (\nodes(t_g))\nonumber  \\
\eea
where $\calJ_{\text{tendon}}$ will be chosen to penalize the control effort, 
and
$\softCost_{\text{approach}}$ and $\softCost_{\text{grasp}}$ can be understood as terminal costs rewarding fingertip configurations at specific times $t_a$ (right before grasping) and $t_g$ (time of grasping) which are assumed to be given.

We approximately solve \eqref{eq:softObjective} by first finding tendon lengths optimizing $\softCost_{\text{approach}} (\nodes(t_a))$ and $\softCost_{\text{grasp}} (\nodes(t_g))$ then solve for intermediate lengths minimizing $\int_{0}^{\finalTime}  \calJ_{\text{tendon}}(\MX^\star, \MY, \restlengths)$.

\myParagraph{Choice of $\calJ_{\text{tendon}}$}
We choose the term $\calJ_{\text{tendon}}$ in the objective function~\eqref{eq:softObjective} as:
\beq
\label{eq:tendonCost}
\calJ_{\text{tendon}}(\MX^\star, \MY, \restlengths) = \max_{0\leq t \leq \finalTime} \left\| \frac{d\restlengths}{dt} \right\|_\infty
\eeq
which penalizes the maximum rate of change of the rest lengths $\restlengths$ (our control inputs) in the interval $[0,\finalTime]$. 
This choice is motivated by the fact that rapid changes in tendon lengths result in large forces on the tendon attachment points. 
These are undesirable, since large forces on specific points (i)  create significant localized deformations which risk damaging the soft gripper, and (ii)  cause large local accelerations that violate the quasi-static assumption. 

\myParagraph{Choice of $\softCost_{\text{grasp}}$}
We choose the term $\softCost_{\text{grasp}}$ in the objective function~\eqref{eq:softObjective} as:
\beq
\label{eq:Cgrasp}
\softCost_{\text{grasp}}(\MY) = \textstyle\sum_{i=1}^4 \| \node_{tip_i} - \vo \|_2^2
\eeq
which encourages the four fingertips $\node_{tip_i} \subset \MY$ to be as close as possible to the target centroid $\vo$ at
the time of grasping.

\begin{figure}
    \begin{center}
    \begin{minipage}{\textwidth}
    \begin{tabular}{cc}%
    \myhspace \hspace{-3mm}
            \begin{minipage}{\mpw}%
            \centering%
            \includegraphics[trim=0mm 0mm 0mm 5mm, clip,height=3.5cm]{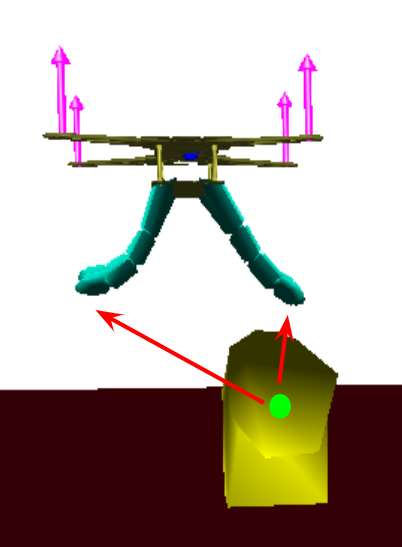} \\
            \hspace{0cm}(a) 
            \end{minipage}
        & \myhspace 
            \begin{minipage}{\mpw}%
            \centering%
            \includegraphics[trim=0mm 0mm 0mm 5mm, clip,height=3.5cm]{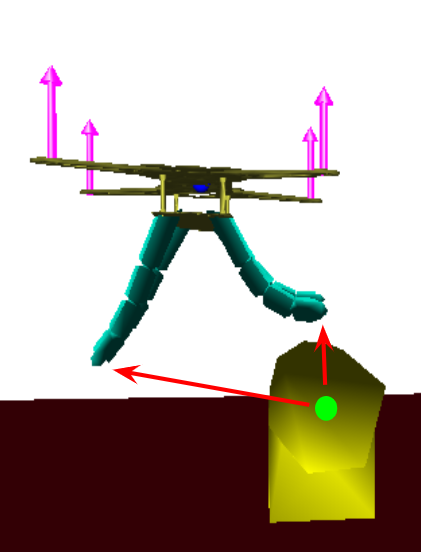} \\
            \hspace{0cm}(b) 
            \end{minipage}
        \end{tabular}
    \end{minipage}
    \vspace{-2mm} 
    \caption{
    {Result of optimizing for two different objectives during the approach phase, based on the vectors (red) from target centroid (green) to fingertips: (a) maximizes (locally) the fingertip distance $\softCost_1$ from the target centroid, and (b) maximizes the norm $\softCost_2$ of the cross product between the vectors connecting the fingertip to the centroid. Since the quadrotor is moving aggressively towards the target, (b) is more likely to yield an enveloping grasp.}
     \label{fig:grasp_objectives} \vspace{-5mm}}
    \end{center}
\end{figure}

\myParagraph{Choice of $\softCost_{\text{approach}}$}
We present two potential choices for the term $\softCost_{\text{approach}}$ in the objective function~\eqref{eq:softObjective}.
The simplest choice is to set $\softCost_{\text{approach}}$ as:
\beq
\label{eq:C1}
\softCost_{1,\text{approach}}(\MY) = - \textstyle\sum_{i=1}^4 \| \node_{tip_i} - \vo \|_2^2
\eeq
which, recalling that minimizing a cost function $f$ is the same as maximizing $-f$, 
simply maximizes the distance between the fingertips $\node_{tip_i}$ and the target centroid $\vo$ at
the time of approach. In the following, we denote this choice as $\softCost_1$.
While optimizing~\eqref{eq:C1} intuitively leads to ``opening'' the fingers of the gripper as much as possible right before grasping
 (hence maximizing the distance to the target centroid, see Fig.~\ref{fig:grasp_objectives}a), such cost may induce suboptimal behaviors, in particular when performing aggressive grasping in which the target is asymmetrically-located with respect to the fingers (i.e. the drone is not directly above the target, see Fig.~\ref{fig:grasp_objectives}).
Therefore, below we consider an alternative cost that 
 rewards the fingertips for surrounding the target so to assure that the target remains between the fingers.

A better choice for the objective for the approach phase of an aggressive grasp is the norm of the cross product of the vectors connecting fingertips and target centroid (Fig.~\ref{fig:grasp_objectives}(b)):
\beq
\label{eq:softCost2}
\softCost_{2,\text{approach}}(\MY) = -\textstyle\sum_{i=1,3} \| (\node_{tip_{i}} - \vo) \times  (\node_{tip_{i+1}} - \vo) \|_2^2
\eeq
 which for simplicity we denote as $\softCost_2$ in the rest of this paper.
Unlike the prior objective, $\softCost_{2}$ rewards both fingertip distance and angle. More rigorously, the cross product denotes the area of the tetrahedron (or triangle in 2D) formed by fingertips and target centroid. All else being equal, the odds of a successful grasp improve with the volume of the target object enveloped by the gripper. Without knowing the target shape, maximization of the volume between fingertips and gripper then implies maximization of the potential target area to be enveloped. 
If we approximate (given finger kinematics and quadrotor motion) that, once the ``grasp'' phase begins, the fingertips will move in a straight line towards the target centroid, then any part of the target object in the area between fingertips and gripper will ultimately be enveloped by the gripper and contribute to a successful grasp.

\subsection{Modeling of a Tendon-Actuated Soft Gripper}
\label{sec:gripperModel}
This section shows how to compute the instantaneous rest length $\restlengths^\star(t_g)$ at time 
$t_g$ such that the corresponding soft gripper configuration minimizes $\softCost_{\text{grasp}}(\MY)$ (the same approach can be used to compute $\restlengths^\star(t_a)$ to minimize $\softCost_{\text{approach}}(\MY)$). 
Note that we can compute an instantaneous rest length thanks to Assumption~\ref{ass:softGripper}, which assumes 
we can neglect the soft gripper dynamics.

\myParagraph{FEM Model}
Our approach follows Bern\setal~\cite{Bern17iros}, using a finite element approximation to compute tendon lengths minimizing an objective $\softCost(\MY)$ (as we mentioned, we will use the same approach to minimize $\softCost_{\text{grasp}}(\MY)$ or $\softCost_{\text{approach}}(\MY)$). 
We contribute analytic expressions for all Jacobians 
\extended{(Appendix~A in~\cite{Fishman20tr-softDrones})}{(Appendix~\ref{subsec:fem})}. We approximate the infinite-dimensional soft gripper configuration $\MY$ as a set of $\nrNodes$ discrete nodes, as in \emph{finite element methods} (FEM).
With slight abuse of notation, we still use $\MY$ to denote the discretized set of nodes:
\begin{equation}
    \MY \doteq [\node_1 \; \node_2 \; \ldots\; \node_n] \in\Real{3\times \nrNodes}
\end{equation}
where $\node_i \in \Real{3}$ is the position of the $i$-th node. 
The nodes are arranged in a \emph{tetrahedral mesh}, and the tendons are approximated as one-sided springs. Finally, a set of pins (also modeled as linear springs) fixes the mesh nodes to the quadrotor base.
Based on this FEM model, we use a Jacobian-based iterative solution to the soft robot inverse kinematics, \ie to find the tendon rest lengths $\restlengths$ that yield a static configuration $\nodes$ minimizing the objective $\softCost(\nodes)$. 

\myParagraph{Inverse Kinematics Overview}
We minimize
$\softCost(\nodes)$ with respect to the rest lengths $\restlengths$ (recall again that $\nodes$ depends on $\restlengths$) 
via gradient descent. 
The complexity in the ``soft case'' is that one cannot write the relation between $\restlengths$ and $\nodes$ analytically. 
To circumvent this issue, and following~\cite{Bern17iros},  
we first solve the forward kinematics problem (determining the system state for a given actuation) by minimizing the total system energy using Newton's method to find quasi-static equilibrium, a system state where net force (and acceleration) are zero. Once an equilibrium configuration is found, we obtain an analytic expression for the actuator Jacobian 
$\frac{d\mathbf{\nodes}}{d\restlengths}$.\footnote{To keep a matrix (rather than a tensor) notation, we assume that all Jacobians involving $\MY$, \eg $\frac{d\mathbf{\nodes}}{d\restlengths}$, work 
on a vectorization of $\MY$.} 
Then we compute the Jacobian $\frac{d\softCost}{d\nodes}$ analytically based on the definition of $\softCost$ in eqs.~\eqref{eq:Cgrasp}-\eqref{eq:softCost2}.
Finally, we use $\frac{d\mathbf{\nodes}}{d\restlengths}$ and $\frac{d\softCost}{d\nodes}$
to compute the gradient of the cost $\frac{d\softCost}{d\restlengths} = \frac{d\softCost}{d\MY} \frac{d\MY}{d\restlengths}$ with respect to the control $\restlengths$ and take a  gradient descent step. The process is iterated till convergence. 
While the computation of $\frac{d\softCost}{d\nodes}$ is straightforward {given $\softCost$},
in the following we describe the expression of the energy and the Jacobian 
$\frac{d\mathbf{\nodes}}{d\restlengths}$. 

\subsection{Jacobian $\frac{d\nodes}{d\restlengths}$ via Forward Kinematics}
We solve the forward kinematics $\restlengths \mapsto \MY$ by minimizing the total energy of a configuration $\MY$ for a given 
choice of $\restlengths$ (as in~\cite{Bern17iros}).
The total energy of the soft gripper can be written as: 
 \begin{equation}
 \label{eq:energy}
\begin{split}
    E(\nodes, \restlengths, \MX) = &E_{mesh}(\nodes) + E_{tendons}(\nodes, \restlengths) \\
                                     + & E_{pins}(\nodes, \MX) + E_{gravity}(\nodes) 
\end{split}
\end{equation}
with the equilibrium configuration $\nodes_{eq}(\restlengths, \MX)$ as the minimizer of the energy:
$\nodes_{eq}(\restlengths, \MX) = \argmin_\nodes E(\nodes, \restlengths, \MX)$.
In the following, we describe each term in the energy~\eqref{eq:energy}. 

\myParagraph{Mesh Energy}
The energy term $E_{mesh}$ models the contribution to the system energy due to deformations of the soft material. 
We compute the energy contribution of each tetrahedral component $\nodes_{ijkl} \doteq [\node_i \; \node_j \; \node_k \; \node_l ]$ separately. We define the relative displacement of each node in the element:
\begin{equation*}
    \defdisp =   
    \begin{bmatrix}
    (\node_i - \node_l) \quad
    (\node_j - \node_l) \quad
    (\node_k - \node_l)
  \end{bmatrix}
\end{equation*}
which contains the relative positions of vertices $i,j,k$ with respect to $l$.
When no forces are applied, the mesh assumes the rest displacement $\restdisp$. 
In the presence of external forces, the mesh assumes a deformed displacement  $\defdisp$.  
The energy of a configuration depends on the mismatch between rest and deformed displacement.
Define the \emph{deformation gradient} $\defgrad\!=\!\defdisp (\restdisp)^{-1}$, 
the rest volume $\restvol \!=\!1/6 \det(\restdisp)$,
\extended{
and the \emph{volumetric deformation} $\voldef \!=\! \det(\defgrad)$.}{
and the \emph{volumetric deformation} $\voldef\!=\!\det(\defgrad)$, which is the ratio of deformed to undeformed volume.}
We use a neo-Hookean material model which defines the mesh energy in terms of Lam\'e parameters $\mu, \lambda$~\cite{Sifakis12siggraph}: 
\begin{equation}
    E^{ijkl}_{mesh}(\nodes) = \restvol\left[ 
    \frac{\mu}{2} tr(\defgrad^T \defgrad - \eye) - \mu \ln (\voldef) + \frac{\kappa}{2} \ln^2(\voldef)
    \right]
\end{equation}
The total mesh energy $E_{mesh}(\nodes) = \sum_{ijkl} E^{ijkl}_{mesh}(\nodes)$ is the sum of the contributions of all elements $ijkl$.

\myParagraph{Tendon Energy}
The energy term $E_{tendon}$ models the contribution of the tendons to the system.
Each tendon is defined by the set of nodes in the mesh it is attached to. 
Let us denote with $i_1,\ldots,i_n$ the set of node indices tendon $i$ is attached to (the so called \emph{routing path}). 
Then, the tendon deformation for tendon $i$ is defined as:
\begin{equation}
\begin{split}
    \tendondef_i &= \textstyle\sum_{k=1}^{n-1}\|\node_{i_{k+1}} - \node_{i_{k}}\|_2 \;-\; \restlengths_i
\end{split}
\end{equation}
which intuitively is the mismatch between the desired routing path length (dictated by the rest length $\restlengths_i$) 
and the actual length according to the mesh nodes ($\sum_{k=1}^{n-1}\|\node_{i_{k+1}} - \node_{i_{k}}\|$).
We can then recover the energy of tendon $i$ by modeling the tendon as a one-sided spring with spring constant $\kappa_{tendon}$:
\begin{equation} 
E^i_{tendon}(\nodes) = 
\left\{
\begin{array}{ll}
0 & \text{if } \tendondef < 0\\
\kappa_{tendon} \; \tendondef_i^2  & \text{otherwise}
\end{array}
\right.
\end{equation} 
The total tendon energy $E_{tendon} = \sum_{i=1}^{8} E^i_{tendon}(\nodes)$ is the sum of the contribution of all tendons.

\myParagraph{Pin Energy}
The energy term $E_{pin}$ models the contribution of the pins (connecting the soft gripper to the quadrotor base) 
to the system.
Each pin $i$ is modeled as a spring with constant $\kappa_{pin}$, connecting a mesh node $\node_i$ belonging to the soft gripper, to a point $\vxx_i^{pin}$  belonging to the quadrotor base (for a given drone state $\MX$). 
The energy for each pin $i$ is:
\begin{equation}
    E^i_{pin}(\node_i, \MX) = \kappa_{pin} \|\node_i - \vxx_i^{pin}\|_2
\end{equation}
The total energy $E_{pin}$ is the sum of the contribution from all pins (we use three pins per finger).

\myParagraph{Gravitational Energy}
The energy term $E_{gravity}$ models the impact on the gravity on the system's energy. 
We approximate the gripper mass as concentrated in the mesh nodes, and denote with $\nodemass_i$ the mass of node $i$.
The gravitational potential energy depends on the mass and height of the node: 
\begin{equation}
    E^i_{gravity}(\nodes) = - \nodemass_i \; \vg\tran \node_i
\end{equation}
where $\vg \doteq [0,0,-9.81]\tran \text{m/s}^2$ is the gravity vector. 
The total gravitational energy $E_{gravity}(\nodes) = \sum_{i=1}^\nrNodes E^i_{gravity}(\nodes)$ is the sum of the contribution of all nodes.

\myParagraph{Jacobian $\frac{d\MY}{d\restlengths}$}
Given a control $\restlengths$ (\eg a point for which we want to obtain a gradient), 
we compute a quasi-static configuration $\MY$ that minimizes the system energy~\eqref{eq:energy} using Newton's method. Then, the actuator Jacobian can be computed from the Hessians of the energy, as shown in~\cite{Bern17iros}:
\beq
\label{eq:jacActuator}
\frac{d\MY}{d\restlengths} = -\frac{d^2E}{d\MY^2}\inv \frac{d^2E}{d\MY d\restlengths}
\eeq
The analytic expressions of the terms on the right-hand-side of~\eqref{eq:jacActuator} is reported in 
\extended{Appendix~A in~\cite{Fishman20tr-softDrones}.}  
{Appendix~\ref{subsec:fem}.} It is worth noting that these terms are readily available as a byproduct of the application of Newton's method to the minimization of~\eqref{eq:energy}.

\subsection{Trajectory Optimization and Open-loop Control for a Tendon-Actuated Soft Gripper}
\label{sec:traj_gripper}

The inverse kinematics model in the previous section allows computing the tendon rest lengths $\restlengths^\star(t_a)$ and $\restlengths^\star(t_g)$ 
that ensure that the fingertips of the soft gripper are 
away from the target at time $t_a$ (in the approach phase) and 
close to the target centroid at time $t_g$ (\ie during the grasp).
With our choice of $\calJ_{\text{tendon}}$ (Eq.~\eqref{eq:tendonCost}) and under Assumption~\ref{ass:softGripper}, solving problem~\eqref{eq:optControlSoftGripper} reduces to 
(i) ensuring that $\restlengths(t_g)$ is equal to $\restlengths^\star(t_g)$ and that $\restlengths(t_a)$ is equal to $\restlengths^\star(t_a)$
and (ii) minimizing the changes of $\restlengths$ in $[0,\finalTime]$. 
It is straightforward to see that the optimal control trajectory under this setup consists in linearly interpolating $\restlengths$ from the initial rest length (at time 0) to the lengths $\restlengths^\star(t_a)$  
(at time $t_a$); then linearly interpolating {}between lengths $\restlengths^\star(t_a)$ to the lengths $\restlengths^\star(t_g)$  
(at time $t_g$), and finally keeping them constant afterwards (until $\finalTime$). 
We apply the resulting control sequence $\restlengths^\star(t)$ in open loop.


\section{Geometric Control and Trajectory Optimization for the Quadrotor Base}
\label{sec:quadrotorBase}

This section describes how to solve the drone control subproblem~\eqref{eq:optControlQuadrotor}. 
Thanks to the decoupling described in Section~\ref{sec:overview}, 
problem \eqref{eq:optControlQuadrotor} falls back to a standard quadrotor control formulation. 
Therefore, as done in related work~\cite{Mellinger11icra}, we solve it by first computing a nominal state trajectory using polynomial trajectory  \extended{optimization}{optimization (briefly reviewed in Section~\ref{sec:traj_quadrotor})} and then using a geometric controller to track the nominal trajectory. 
The element that sets our setup apart is the presence of (the disturbance) $\MY$ in the quadrotor dynamics in~\eqref{eq:optControlQuadrotor}.
{Intuitively, the soft gripper in general imposes a torque that acts to orient the quadrotor towards level.
While this torque may prevent the achievement of the control goals or even destabilize the platform}, 
in the following we show that under certain assumptions on the soft load (Fig.~\ref{fig:geoControl}) and the aerodynamic forces experienced by the quadrotor, we can bound the disturbance torque such that a standard geometric controller remains asymptotically stable. 
{In practice we empirically observe that our platform remains stable even when these assumptions are violated (see experiments in Section \ref{sec:experiments}).

\subsection{Minimum-Snap Trajectory Optimization}
\label{sec:traj_quadrotor}

We first compute a nominal trajectory (quadrotor state and its derivatives over time) 
by solving~\eqref{eq:optControlQuadrotor} and neglecting the presence of the disturbance $\MY$.  
As done in related work~\cite{Mellinger11icra}, (i) we consider a cost function in~\eqref{eq:optControlQuadrotor} that penalizes the integral of the 4th derivative of the state (minimum snap), (ii) we assume polynomial trajectories, and (iii) we leverage differential 
flatness to express the optimal control problem as 4 decoupled scalar optimization problems over the flat outputs
(three for the Cartesian position of the quadrotor and one for its yaw angle).
Mellinger\setal~\cite{Mellinger11icra} and Bry\setal~\cite{Bry15ijrr-aggressiveFlight} 
show that the resulting polynomial optimization problems can be solved efficiently via quadratic programming.

\subsection{Geometric Control of a Quadrotor with a Soft Load}
\label{sec:control_quadrotor}
Given the quadrotor trajectory $\MX^\star(t)$ generated according to Section~\ref{sec:traj_quadrotor}, 
we are only left to design a controller that is able to track  $\MX^\star(t)$ in the face of external disturbances, including 
the torque induced by the time-varying soft payload $\MY$. 
Here we show that a commonly adopted solution, the geometric controller by Lee\setal~\cite{Lee10cdc-geometricControl}, preserves asymptotic stability 
even in the presence of our soft gripper.
We first review the basics of the geometric controller, {then discuss its stability with the added soft load}. 

\begin{figure}[t]
    \begin{center}
    \begin{minipage}{\textwidth}
    \begin{tabular}{cc}%
    \myhspace \hspace{-7mm}
            \begin{minipage}{\mpw}%
            \centering%
            \includegraphics[trim=0mm 0mm 0mm 0mm, clip,height=3.2cm]{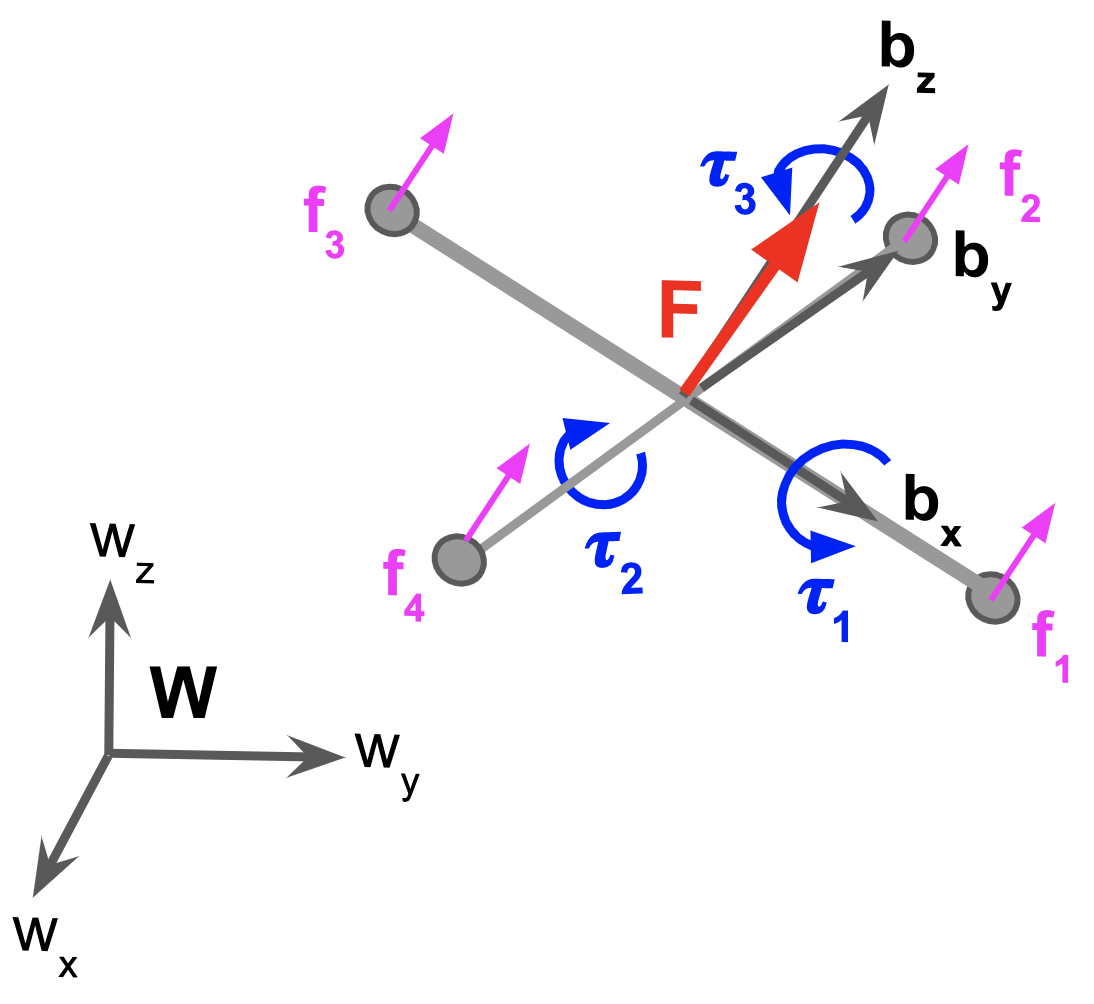} \vspace{-6mm}\\
            \hspace{0cm}(a) 
            \end{minipage}
        & \hspace{-9mm}
            \begin{minipage}{\mpw}%
            \centering%
            \includegraphics[trim=0mm 0mm 0mm 0mm, clip,height=2.5cm]{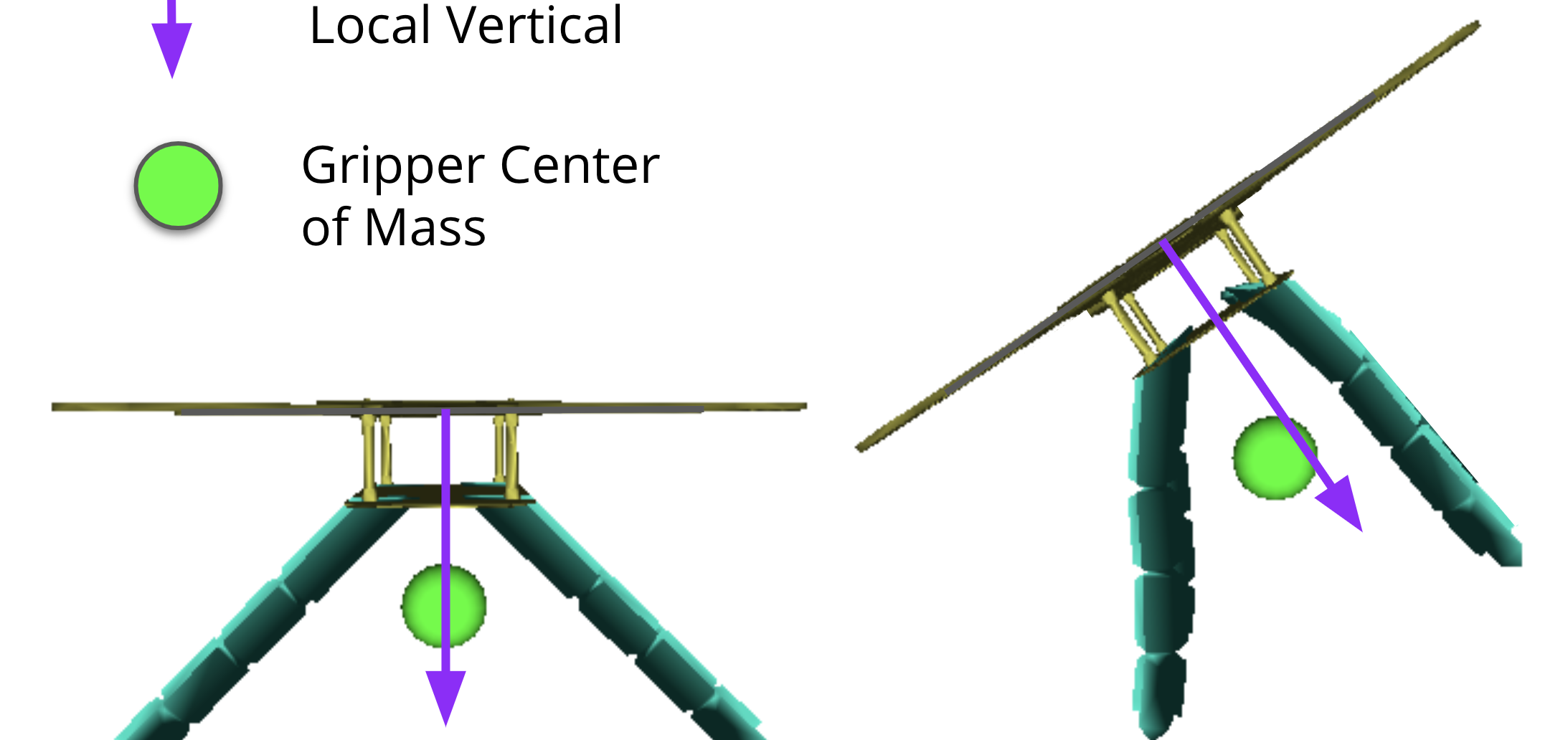} \\
            \hspace{0cm}(b) 
            \end{minipage}
        \end{tabular}
    \end{minipage}
    \begin{minipage}{\textwidth}
    \end{minipage}
    \vspace{-2mm} 
    \caption{
    {(a) Quadrotor forces, torques, local and world frames.
    (b) {Without external perturbation, the soft gripper center of mass is aligned with the local vertical due to the symmetry of its fingers; forces that deform the gripper might cause its center of mass to be misaligned.}}
    \label{fig:geoControl}\vspace{-7mm} }
    \end{center}
\end{figure}

\myParagraph{Geometric Controller}
In the following, we explicitly write the quadrotor state (that we generically denoted with $\MX$ so far) 
as $\MX \doteq \{ \quadpos, \quadrot, \dot{\quadpos}, \quadrvel \}$, including the 
quadrotor position $\quadpos \in \Real{3}$, 
its rotation $\quadrot\in \SOthree$,
the linear velocity $\dot{\quadpos} \in \Real{3}$, 
and the angular velocity $\quadrvel \in \Real{3}$.
 Using this notation, and denoting the columns of $\quadrot$ as $\quadrot = [\rotcolx \; \rotcoly \; \rotcolz]$,  
the quadrotor dynamics can be written as: 
\begin{equation}
\left\{
\begin{array}{ll}
    m \Ddot{\quadpos} = m \vg + f \rotcolz \\
    \dot{\quadrot} = \quadrot \hat{\quadrvel} \\
    \MJ \dot{\quadrvel} = - \quadrvel \times \MJ \quadrvel + \vtau+ \vtau_{load}\\
\end{array}
\label{eq:quadrotor_dynamics}
\right.
\end{equation}
where $m$ is the total mass of the platform and gripper, 
$\vg$ is the gravity vector, 
$\MJ$ is the moment of inertia, 
$f$ is the scalar thrust force (applied at the quadrotor center of mass and along the local vertical direction $\rotcolz$) 
resulting from the propeller forces $\propforces_1,\propforces_2,\propforces_3,\propforces_4$, 
$\vtau \in \Real{3}$ is the torque resulting from the propeller forces, 
and $\vtau_{load}$ is the torque exerted by the soft gripper (Fig.~\ref{fig:geoControl}(a)).
In~\eqref{eq:quadrotor_dynamics}, the symbol $\times$ is the vector cross product, 
the \textit{hat} map $\hat{\cdot}$ maps a 3D vector to a $3\!\times\!3$ skew symmetric matrix
and its inverse, the \textit{vee} map $\cdot^{\vee}$, maps a $3\!\times\!3$ skew-symmetric matrix to a vector 
(as in~\cite{Lee10cdc-geometricControl}). 

The geometric controller~\cite{Lee10cdc-geometricControl} takes as input a desired state 
$\MX^\star \doteq \{ \quadpos_d, \quadrot_d, \dot{\quadpos}_d, \quadrvel_d \}$,\footnote{In~\cite{Lee10cdc-geometricControl}, 
the desired  rotation $\quadrot_d$ and angular velocity $\quadrvel_d$ are built from a desired yaw angle. 
We refer the reader to~\cite{Lee10cdc-geometricControl} for details.} 
and computes the tracking errors:
\begin{equation}
\hspace{-1mm}
\label{eq:trackingErrors}
\begin{array}{ll}
\ve_p = \quadpos - \quadpos_d,  & \text{(position error)} \\
\ve_v = \Dot{\quadpos} - \Dot{\quadpos}_d & \text{(linear velocity error)} \\
\ve_r = \frac{1}{2}(\quadrot_d\tran \quadrot - \quadrot\tran \quadrot_d)^{\vee} & \text{(rotation error)} \\
\ve_{\Omega} = \quadrvel - \quadrot\tran \quadrot_d \quadrvel_d & \text{(angular velocity error)}
\end{array} 
\end{equation}
Then the controller decides for suitable thrust force $f$ and torques $\vtau$ to 
contrast these errors using the control law:
\bea
\label{eq:geoController}
 f &=& -\rotcolz\tran (k_p \ve_p + k_v \ve_v + m \vg - m \Ddot{\quadpos}_d)  \\
    \vtau &=& -k_r \ve_r - k_{\Omega} \ve_{\Omega} + \quadrvel \times \MJ \quadrvel \\
          & & - \MJ( \hat{\quadrvel} \quadrot\tran \quadrot_d \quadrvel_d - \quadrot\tran \quadrot_d \dot{\quadrvel}_d     )
\eea
where $k_p, k_v, k_r, k_\Omega$ are suitable control gains. 
We refer the reader to~\cite{Lee10cdc-geometricControl} for details about how to 
map the desired total thrust $f$ and torque $\vtau$ 
to propeller forces $\propforces_1,\propforces_2,\propforces_3,\propforces_4$.
\myParagraph{Asymptotic Stability} 
Here we show that the geometric controller above is stable despite the presence of the soft gripper. We assume that the quadrotor is upright, i.e. $[0 \; 0 \; 1] \cdot \rotcolz \geq 0$.
The key insight is that as it deforms due to gravity and the quadrotor acceleration,\footnote{{Note that in the absence of aerodynamic drag, the torque on the gripper due to gravity is equal and opposite to the torque due to the quadrotor acceleration; gravity as a uniform body force is unable to exert a net torque on the system. In this case, the quasi-static load does not deform (remains aligned with the local vertical) and exerts no torque, and the system stability is unaffected by the load. As aerodynamic forces affect the quadrotor acceleration, the gripper deforms as we discuss.}} 
{the soft gripper
center of mass remains below the local vertical as long as the drone's horizontal thrust and velocity are aligned\footnote{{The opposite case -- in which thrust and aerodynamic drag are both acting opposite the quadrotor's horizontal velocity -- is of necessity transient, as both drag and thrust are acting to restore the alignment of horizontal velocity with horizontal thrust. See Appendix~\ref{subsec:controller} for further discussion.}}} (Fig.~\ref{fig:geoControl}(b)); 
this implies that the soft load imposes a torque $\vtau_{load}$ which acts to orient the quadrotor towards level. 
This observation, associated with the assumption that the desired final state is level and that the soft gripper is symmetric (which ensures that no torque is asserted when level), 
allows proving the following result.

\begin{theorem}[Stability of Velocity and Attitude Controller] $\text{ }$ $\text{ }$
\label{thm:controllerConvergence} 
Consider a quadrotor confined to the vertical x-z plane~\cite{Thomas14bioinspiration}, 
with a {soft payload which is symmetric about the quadrotor's thrust axis when no external forces are applied (Fig.~\ref{fig:geoControl})}.
Denote with $\loadml$ the load's first moment of mass along the $\rotcolz$ axis when the quadrotor is level.
Assume that $k_x = 0$ (no position control) and $\Ddot{\quadpos}_d = \zero$ (no desired acceleration, so the quadrotor is level at the desired state), and that 
\begin{equation}
\begin{split}
    \rotcolz \cdot [0, 0, 1] &\geq 0 \\
    |\vtau_{load}| &\leq |\loadml  \rotcolz \times \vg|\\
    \sign (\vtau_{load}) &= \sign(\rotcolz \times \vg)\\
\end{split}
\label{eq:quad_assumptions_1}
\end{equation}
Then the geometric controller in~\eqref{eq:geoController}
asymptotically stabilizes the quadrotor velocity and attitude in~\eqref{eq:quadrotor_dynamics}. 
\end{theorem}

The proof is given in 
\extended{Appendix~B in~\cite{Fishman20tr-softDrones}.} 
{Appendix~\ref{subsec:controller}, which also provides a more extensive discussion about the assumptions in~eq.~\eqref{eq:quad_assumptions_1}.
Theorem~\ref{thm:controllerConvergence}
proves convergence for the attitude and velocity controller. However, in the next section, we additionally demonstrate experimental convergence of the position controller, as well as successful performance during agile grasping -- which in practice entails relaxing the requirement that the quadrotor thrust and velocity align and that the soft gripper remains symmetric with respect to the quadrotor vertical}.

\subsection{Summary: Grasp Planning for a Soft Aerial Manipulator}
Our complete planning/control approach is as follows.
First, we plan a minimum-snap polynomial trajectory towards the target with a predefined grasp time $t_g$ and fixed velocity at the moment of grasp.
Given this trajectory, we use root-finding to find the time at which the horizontal distance from quadrotor to target is small enough that the base of the leading fingertips is above the target. 
We choose this time as the end of the approach phase $t_a$ and compute optimal tendon lengths given this quadrotor position; similarly, we compute optimal tendon lengths for time $t_g$ (Section~\ref{sec:traj_gripper}). 
Then, at execution time, 
the quadrotor will track the minimum-snap polynomial trajectory  using a globally-stable geometric controller, which converges despite the addition of the soft gripper (Section~\ref{sec:control_quadrotor}). 
At the same time, we will execute the control actions for the soft gripper in open loop, interpolating from the initial tendon lengths to the optimal lengths at $t_a$ and $t_g$.  
Using this methodology we consistently achieve grasps at speeds up to 0.5m/s.


\section{Experiments}
\label{sec:experiments}

We validate our soft aerial manipulator design
in \emph{SOFA}~\cite{Faure12sofaam}, a popular open-source soft dynamics simulator with dedicated plugins for tendon-actuated soft manipulators \cite{Duriez13icra}. 
The experiments show that 
(i) the geometric controller converges regardless of the soft payload, 
(ii) the platform can reliably grasp objects of unknown shape, and 
(iii) the decoupled controller is amenable for real-time execution. 

\subsection{Setup}
\label{sec:exp_setup}

We simulate our soft aerial manipulator in SOFA~\cite{Faure12sofaam} (see Fig.~\ref{fig:timelapse} 
and the {video attachment}). 
We choose a simulation timestep of 0.01 seconds.
The rigid frame of the manipulator is modeled 
after the frame of the \emph{Intel Ready to Fly quadrotor} (size: $0.25\times0.25\times0.04$\SI{}{m}), 
while the four fingers are modeled as described in Section~\ref{sec:softGripper} (each with 
size: $0.18 \times 0.025 \times 0.025\SI{}{m}$).
We choose quadrotor mass $m = \SI{1}{kg}$, inertia $\MJ = \diag{[0.08, 0.08, 0.14]}
\SI{}{kg \cdot m^2}$ and body drag coefficient $0.5$ (applied on the quadrotor centroid). 
As material parameters,  
we choose Young's modulus $\youngModulus=2\cdot 10^4 \SI{}{N/m^2}$ (similar to silicone rubber) 
and Poisson's ratio $\poissonRatio=0.25$, 
and derive Lam\'e parameters as $\mu = \frac{\youngModulus}{ 2 (1 + \poissonRatio)} = 8000 \SI{}{N/m^2}, 
\kappa=\frac{\poissonRatio \youngModulus}{(1 + \poissonRatio) (1 - 2\poissonRatio)} = 6667 \SI{}{N/m^2}$~\cite{Sifakis12siggraph}.
We choose a gripper density $\rho = 250 \SI{}{kg/m^3}$.
The controller gains are set to  $k_p = 16, k_v = 5.6, k_r = 8.81, k_{\Omega} = 2.54$, as in~\cite{Lee10cdc-geometricControl}. 

\subsection{Geometric Control Evaluation}
\label{sec:exp_control}

Fig.~\ref{fig:geometric_error_initializations} plots the norm of the velocity, position, and rotation tracking errors 
defined in~\eqref{eq:trackingErrors} for 20 runs of the geometric controller. In each run, we 
chose a random target location on the {circle of radius $1\SI{}{m}$} (similar to~\cite{Thomas14bioinspiration}, \cite{Cabellero18iros}).
The figure shows quick convergence to the desired state, with position error decreasing by 95\% within {1.3\SI{}{s}}.
The shaded area shows the 1-sigma standard deviation for the tracking errors. 

\begin{figure}
    \centering
    \includegraphics[width=.52\textwidth]{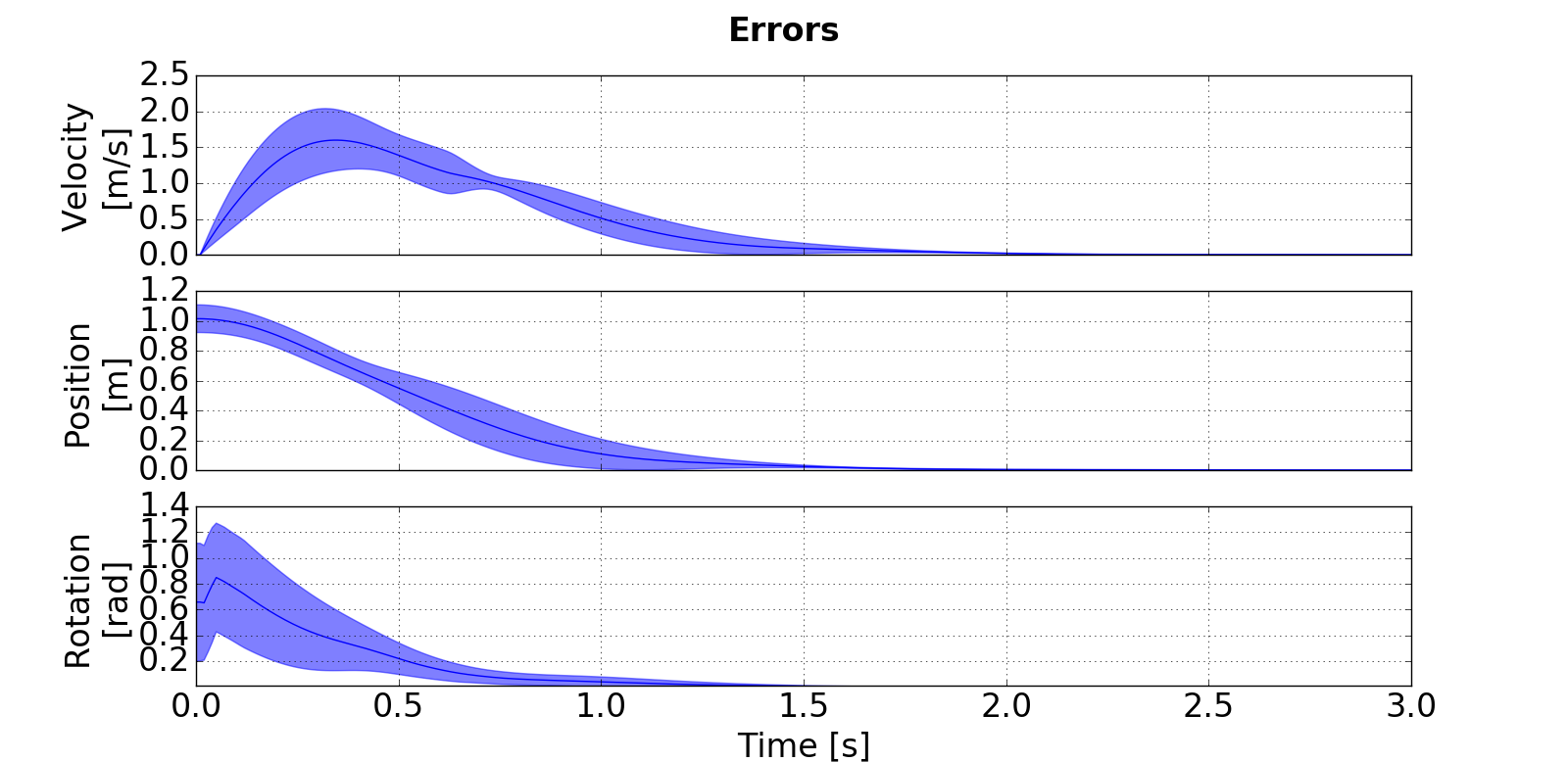}
    \caption{Mean and standard deviation of the tracking errors with 
    gripper density $\rho = 250 \SI{}{kg/m^3}$. Statistics are computed over 20 runs with 
    randomly chosen target locations {on the unit circle}. 
    \label{fig:geometric_error_initializations}  \vspace{-1mm}}
\end{figure}

Fig.~\ref{fig:geometric_error_densities} shows that convergence occurs regardless of the disturbance 
induced by the gripper mass. 
We simulate increasing gripper densities 
$\rho = \{10^{-2}, 10^3, 10^5\}  \SI{}{kg/m^3}$, ranging from a gripper ten times lighter than helium 
to one five times denser than lead; for each density, we repeat 20 runs and plot the 
tracking errors in Fig.~\ref{fig:geometric_error_densities}.  
The figure shows that, while the increased gripper density impacts the convergence rate (in particular, larger densities 
induce an increased overshoot and longer convergence tails), the controller is still able to 
{converge to the 
desired state within $5\SI{}{s}$}. 

\begin{figure}
\centering
\includegraphics[width=.52\textwidth]{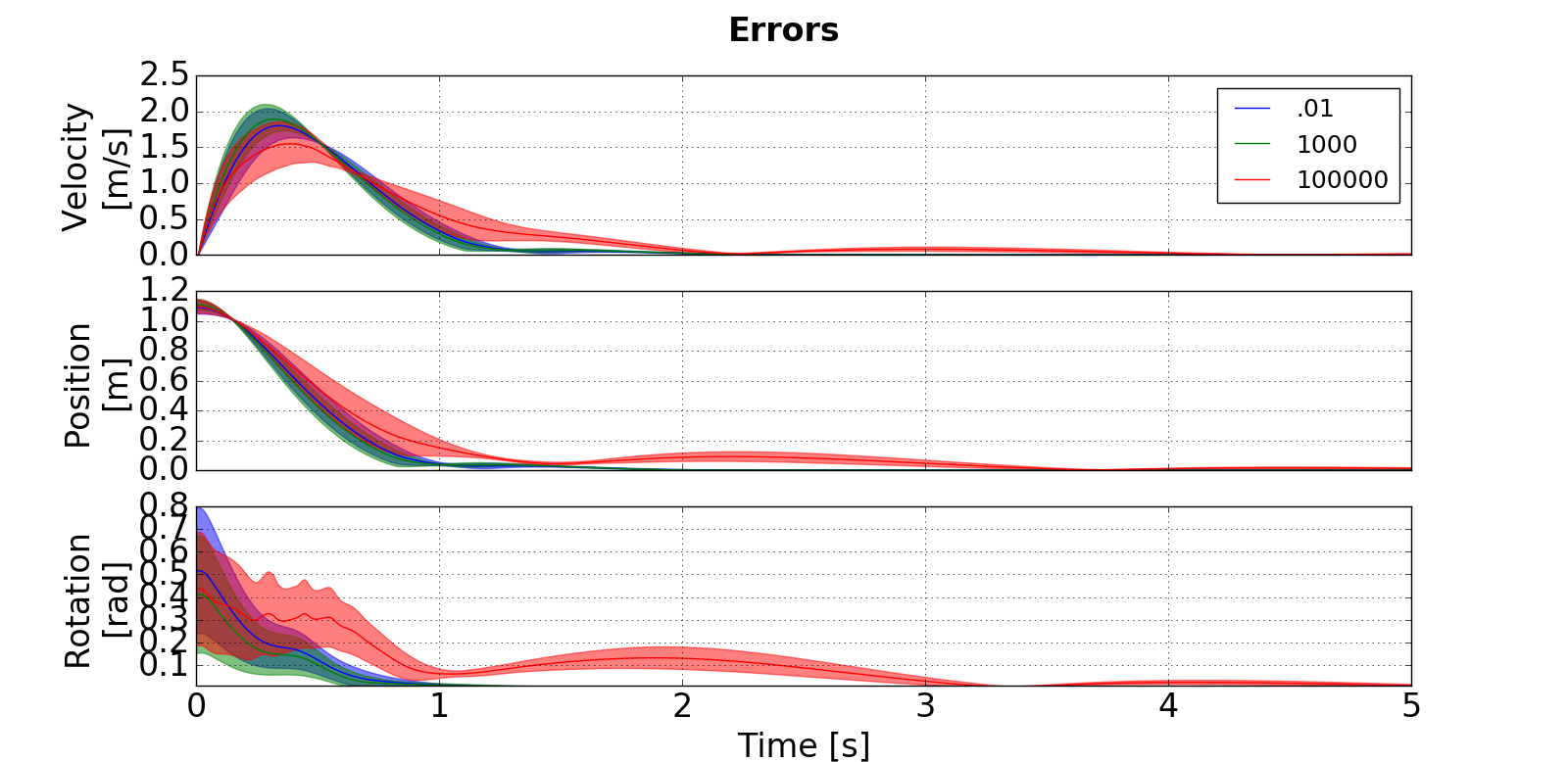}
\caption{Mean and standard deviation of the tracking errors for gripper densities $\rho = \{10^{-2}, 10^3, 10^5\}  
\SI{}{kg/m^3}$.\label{fig:geometric_error_densities} \vspace{-1mm}}
\end{figure}

\subsection{Aerial Grasping Experiments}
\label{sec:exp_grasp}

To validate the effectiveness of the proposed soft aerial  manipulator, we task it to grasp a target of unknown shape and mass. In the simulator, we set the target mass to $\SI{0.05}{kg}$ 

Fig.~\ref{fig:grasp_outcomes1}(a) shows the binary grasp outcome (success/failure) as a function 
of the initial quadrotor height (``z'') and horizontal position (``x'') with respect to the target {using objective $\softCost_{1}$ to compute tendon lengths during the approach phase. For this objective, we require that the horizontal velocity be zero at the moment of grasp and add a trajectory setpoint $0.2\SI{}{m}$ above the target}.
Our proposed soft gripper is able to successfully grasp for all initial conditions with $z > \SI{0.25}{m}$, 
corresponding to the cases where the fingertips start above the height of the target.
To put things in perspective, Fig.~\ref{fig:grasp_outcomes1}(b) shows
the same statistics for a more ``rigid'' design, where we chose Young's modulus to be 
$\youngModulus=2\cdot 10^5 \SI{}{N/m^2}$ (10 times stiffer that our design, with a Young's modulus analogous to a stiff rubber). Comparing Fig.~\ref{fig:grasp_outcomes1}(a)-(b) we realize that the stiffer gripper is more likely to fail. In particular, the stiffer gripper failed in all conditions with $x > 0.3$, in which too much momentum was transferred to the target, hence preventing a successful grasp. 

\begin{figure}
    \centering
    \includegraphics[width=.48\textwidth]{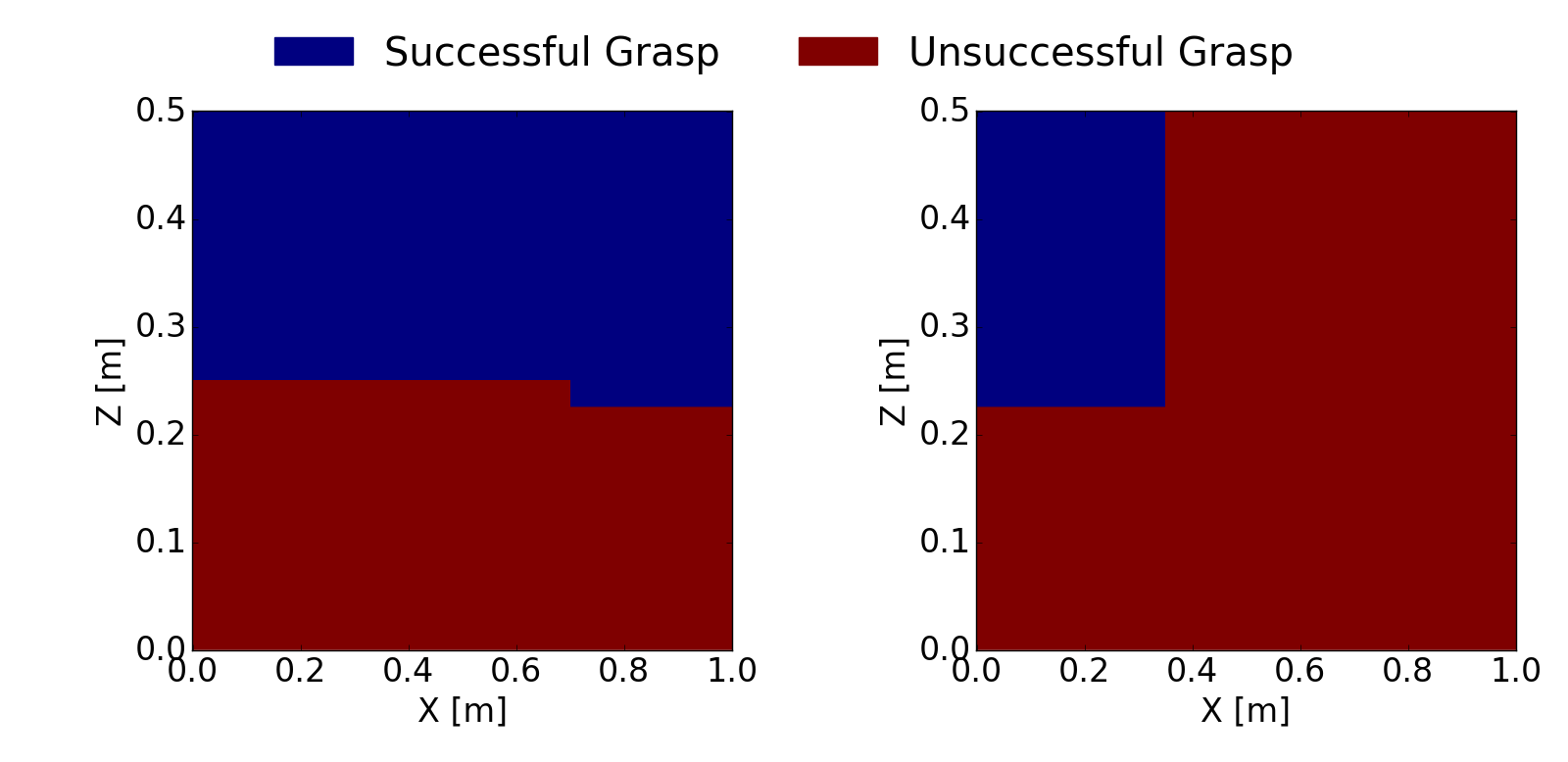}
    \caption{Grasp outcome as a function 
    of the initial quadrotor height (``z'') and horizontal position (``x'') with respect to the target {using objective $\softCost_1$ in the approach phase and with zero horizontal velocity at grasp}. 
    (a) proposed soft aerial manipulator; (b) more rigid design with higher Young's modulus  
   ($\youngModulus=2\cdot 10^5 \SI{}{N/m^2}$).
   \label{fig:grasp_outcomes1} \vspace{-1mm}}
\end{figure}

\begin{figure}
    \centering
    \includegraphics[width=.48\textwidth]{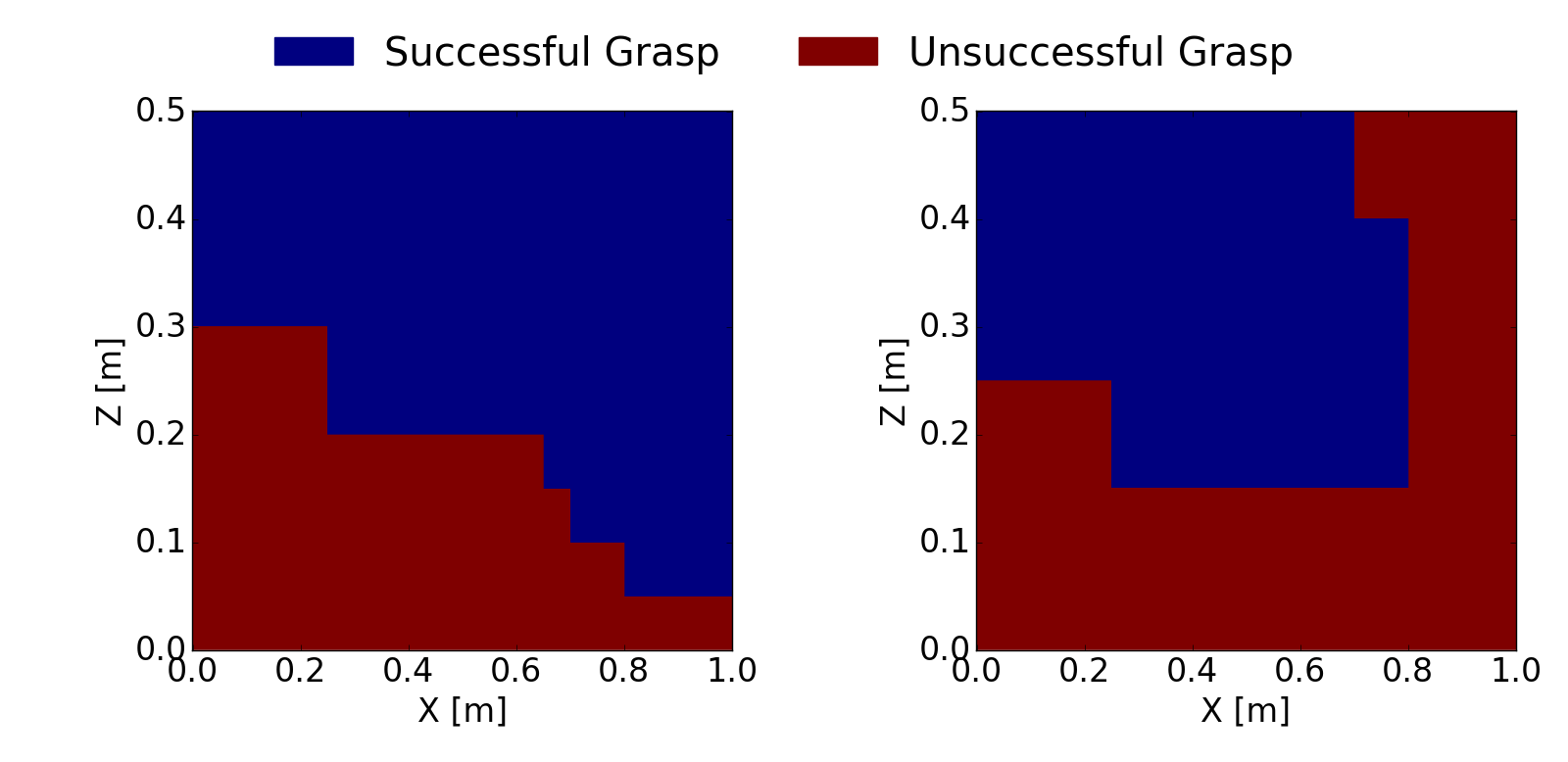}
    \caption{{Grasp outcome as a function 
    of the initial quadrotor height (``z'') and horizontal position (``x'') with respect to the target using objective $\softCost_2$ in the approach phase and with velocity at grasp of $[0.5, 0, -0.125]~\SI{}{m/s}$.
    (a) proposed soft aerial manipulator; (b) more rigid design with higher Young's modulus  
   ($\youngModulus=2\cdot 10^5 \SI{}{N/m^2}$).}
   \label{fig:grasp_outcomes2} \vspace{-4mm}}
\end{figure}

{We carried out a similar test using the aggressive grasp objective $\softCost_{2}$  to compute tendon lengths during the approach phase, a fixed velocity of $[0.5, 0, -0.125]~\SI{}{m/s}$ at the moment of grasp (Fig.~\ref{fig:grasp_outcomes2}) and a $\SI{0.01}{kg}$ target. As above, our proposed gripper design succeeded in all cases when the fingertips can pass above the target (which is easier when quadrotor and target are farther apart), while the stiffer gripper ($\youngModulus=2\cdot 10^5 \SI{}{N/m^2}$) imparts too much momentum to the target when starting farther away.}

These tests exemplify the advantages of softness in aerial manipulation. 
The softer gripper adapts to deviations from the nominal quadrotor trajectory and 
naturally 
mitigates the impact of contact forces on the quadrotor and the target,
thus enabling successful grasps from a wide range of initial conditions in which a more rigid solution fails.

\subsection{Timing}
\label{sec:exp_timing}

{Our C++ implementation of the soft gripper trajectory optimization approach in Section~\ref{sec:softGripper} requires $\approx\SI{1}{s}$ to compute a control sequence for the tendon rest lengths on an Intel Core i7-5500U CPU.}
Note that this can be computed offline before execution. The control is executed in open-loop, hence the computational cost to control the soft gripper during execution is negligible: interpolating the tendon actuations as discussed in~Section~\ref{sec:traj_gripper} requires less than a millisecond.

For the minimum-snap quadrotor trajectory optimization, we use the \emph{cvxopt} package in python.
Our code requires $\approx\SI{0.02}{s}$ to compute a minimum-snap trajectory (as before, this can be done offline). 
The implementation of the geometric controller is also in python and it requires 
$\approx\SI{0.01}{s}$ to compute the instantaneous control action to be applied to the quadrotor propellers.
In summary, the total computation required during execution is in the order of tens of milliseconds, and 
 can be further 
reduced via an optimized multi-threaded C++ implementation.

\section{Conclusion}
\label{sec:conclusion}
We presented a soft aerial manipulator that fully exploits compliance to enable aggressive grasping of unknown objects. 
We proposed a decoupled control and trajectory optimization approach for the soft gripper and the rigid quadrotor base, 
and showed theoretically and experimentally that the quadrotor is stable despite disturbance from the soft gripper. Finally, we observed that our system achieves consistent grasp of a target object in realistic simulations, and 
 is a promising alternative to a more rigid design.  

Future work includes real implementation and testing of the soft aerial manipulator proposed in this paper.
 More fundamentally, we also plan to investigate the proposed ``soft drone'' design for other applications, including 
 aggressive landing, perching, and collision-resistant navigation. 

\section*{Acknowledgments}
The authors gratefully acknowledge Vibha Agarwal for her contribution to the literature review of this paper, 
Mostafa Mohammadi for the useful suggestions on an earlier draft of this paper,
and Philipp F{\"o}hn for a relevant discussion of the torques experienced by a quadrotor load.

\extended{
	{
		\tiny 
		\bibliographystyle{IEEEtran}
		\bibliography{./references/refs,myRefs} %
	}
}
{
	\appendices{}

\section{Finite Element Methods for Soft Manipulators}
\label{subsec:fem}

This appendix provides an extended description of the finite element model (FEM) of our soft gripper,
 including details on how to compute the energy of the soft gripper (Sections~\ref{app:fem-background}-\ref{app:fem-gravity}) 
 and how to obtain the actuator Jacobian (Section~\ref{app:fem-jacobian}) required to implement the inverse kinematics of Section~\ref{sec:softGripper} in the main manuscript. 

\subsection{FEM Model and Energy}
\label{app:fem-background}

A standard approach to model a continuously deformable body is to discretize it into  a finite set of nodes~\cite{Sifakis12siggraph}:
\begin{equation}
    \nodes \doteq [\node_1 \; \node_2 \; \ldots\; \node_n] \in\Real{3\times \nrNodes}
\end{equation}
where $\node_i \in \Real{3}$ is the position of the $i$-th node.

These nodes are organized in a tetrahedral mesh, where each mesh element includes four not necessarily unique nodes.
In our soft aerial manipulator (following Bern\setal~\cite{Bern17iros}),
the mechanical elements are described by this mesh (parametrized by Lam\'e parameters $\mu$ and $\kappa$), tendons (defined by a list of nodes the tendon passes through, rest length $\restlength$ and stiffness $\kappa_{tendon}$) and pins (defined by a Cartesian position and stiffness $\kappa_{pin}$).

The total energy of the soft body is the collective contribution of all of these components as well as gravity. This depends on the node positions $\nodes$, tendon rest lengths $\restlengths$ and quadrotor position $\MX$.
Mathematically, the total energy can be written as: 
\begin{equation}
\label{eq:total_energy}
\begin{split}
    E(\nodes, \restlengths, \MX) = &E_{mesh}(\nodes) + E_{tendons}(\nodes, \restlengths) \\
                                     + & E_{pins}(\nodes, \MX) + E_{gravity}(\nodes) 
\end{split}
\end{equation}

The \emph{force} acting on each node is the negative gradient of the total energy with respect to the node position (a matrix of size $3 \times \nrNodes$), and the \emph{stiffness} is the Jacobian of the force or the negative Hessian of the energy; 
in our derivation, we vectorize $\nodes$ to obtain a Hessian of size $3 \nrNodes \times 3 \nrNodes$. All Hessians or Hessian components are vectorized or referred to specifically using component-wise notation.

In the following we derive analytic expressions for all energies, forces, and stiffnesses associated with our mesh model. All of these are required to compute a stable (energy-minimizing) configuration with Newton's Method. We additionally describe an expression for the actuator Jacobian $\frac{d\nodes}{d\restlengths}$, the mapping from changes in tendon rest lengths to changes in node positions at a stable configuration (following \cite{Bern17iros}), which is computed from forces and stiffness already calculated in the process of energy minimization.

\subsection{Mesh \Properties}
\label{app:fem-mesh}

Our FEM mesh analysis follows~\cite{Sifakis12siggraph}.

The energy contribution of each tetrahedral element $\nodes_{ijkl} \doteq [\node_i \; \node_j \; \node_k \; \node_l ]$ is computed independently. We define the relative displacement of each node in the element:
\begin{equation*}
    \defdisp =   
    \begin{bmatrix}
    (\node_i - \node_l) \quad
    (\node_j - \node_l) \quad
    (\node_k - \node_l)
  \end{bmatrix}
\end{equation*}
which contains the relative positions of vertices $i,j,k$ with respect to vertex $l$.

When no force is applied, the element assumes the rest displacement $\restdisp$; otherwise it assumes a deformed displacement $\defdisp$. These define the deformation gradient $\defgrad$, which is a linearised map between rest and deformed configurations:
\begin{equation}
    \defgrad = \defdisp (\restdisp)^{-1}
\end{equation}
The rest volume $\restvol$ of the element is:
\begin{equation}
    \restvol = \frac{1}{6}|\det \restdisp|
\end{equation}
Following~\cite{Bern17iros} we use a Neo-Hookean material model, in which the strain energy density of each element is defined in terms of $\defgrad$ and $\voldef=\det(\defgrad)$, the ratio of deformed to undeformed volume of the element:
\begin{equation}
\begin{split}
     \psi &= \frac{\mu}{2} tr(\defgrad\tran \defgrad - \eye) - \mu \ln (\voldef) + \frac{\kappa}{2} \ln^2(\voldef) \\
\end{split}
\end{equation}
and energy $E^{ijkl}_{mesh} = \psi \restvol$, where $\mu$ and $\kappa$ are the Lam\'e parameters. 
The volumetric component of the energy $- \mu \ln (\voldef) + \frac{\kappa}{2} \ln^2(\voldef)$ is only quasi-convex in general; however, it is convex for $\voldef < \e^{\frac{\mu}{\kappa} + 1}$ with $\frac{\mu}{\kappa} \geq 0$, so that for any material properties convexity is guaranteed when the ratio of deformed to undeformed volume is less than $\e$.

The nodal force, or the negative energy gradient, is defined in terms of the First Piola Stress Tensor $\piola$:
\begin{equation}
    \piola = \mu(\defgrad - \defgrad^{-T}) + \text{$\kappa$} \ln(\voldef) \defgrad^{-T}
\label{eq:piola}
\end{equation}
The force on the first three nodes in the element is:
\begin{equation}
    \begin{bmatrix} \meshforce^i & \meshforce^j & \meshforce^k \end{bmatrix} = -\restvol \piola \restdisp^{-T}
\label{eq:meshforce}
\end{equation}
and by conservation of momentum $\meshforce^l = -(\meshforce^i + \meshforce^j + \meshforce^k)$.

The mesh stiffness is the gradient of nodal forces with respect to their positions. For the purpose of this calculation we vectorize the element $\nodes_{ijkl} \in \Real{12}$. Let ${\idx} = 1,2,\ldots,12$ be an index and $\nodes_{ijkl}\at{\idx}$ be the $\idx$-th entry of $\nodes_{ijkl}$. First, we calculate the gradient for the stress tensor, $\frac{d \piola}{d \nodes_{ijkl}}$. We define $\frac{d \defdisp}{d \nodes_{ijkl}\at{\idx}}$ as the constant $3 \times 3$ matrix representing the gradient of the deformed displacement matrix with respect to coordinate $\nodes_{ijkl}\at{\idx}$. We first compute the gradient of the deformation gradient $\defgrad$ with respect to coordinate $\nodes_{ijkl}\at{\idx}$: 
\begin{equation}
\begin{split}
    \frac{d\defgrad}{d\nodes_{ijkl}\at{\idx}} &=  \frac{d \defdisp}{d \nodes_{ijkl}\at{\idx}} \restdisp^{-1} \\
\end{split}
\label{eq:dGdY}
\end{equation}
We use $\frac{d\defgrad}{d\nodes_{ijkl}\at{\idx}}$ in~\eqref{eq:dGdY} to compute the gradient of the stress tensor $\piola$~\eqref{eq:piola} with respect to coordinate $\nodes_{ijkl}\at{\idx}$:
\begin{equation}
\begin{split}
    \frac{d \piola}{d \nodes_{ijkl}\at{\idx}} &= (-\mu + \kappa \ln(\voldef)) 
    \left(-\defgrad^{-1} \frac{d\defgrad}{d\nodes_{ijkl}\at{\idx}} \defgrad^{-1} \right)\tran \\
    & \quad + \kappa \ \trace{ 
    \defgrad^{-T} \frac{d\defgrad}{d\nodes_{ijkl}\at{\idx}} }  \defgrad^{-T} + \mu \frac{d\defgrad}{d\nodes_{ijkl}\at{\idx}} \\
\end{split}
\end{equation}
where $\frac{d \piola}{d \nodes_{ijkl}\at{\idx}}$ is a $3 \times 3$ matrix. The relationship of nodal stiffness to the gradient of the stress tensor is the same as the relationship of the nodal forces to the stress tensor \eqref{eq:meshforce}:
\begin{equation}
\begin{split}
    \begin{bmatrix} \frac{d\meshforce^i}{d\nodes_{ijkl}\at{\idx}} & \frac{d\meshforce^j}{d\nodes_{ijkl}\at{\idx}} & \frac{d\meshforce^k}{d\nodes_{ijkl}\at{\idx}} \end{bmatrix} &= -\restvol \frac{d \piola}{d\nodes_{ijkl}\at{\idx}} \restdisp^{-T} \\
    -\left(\frac{d\meshforce^i}{d\nodes_{ijkl}\at{\idx}} + \frac{d\meshforce^j}{d\nodes_{ijkl}\at{\idx}} + \frac{d\meshforce^k}{d\nodes_{ijkl}\at{\idx}} \right) &= \frac{d\meshforce^l}{d\nodes_{ijkl}\at{\idx}} 
\end{split}
\end{equation}
We vertically concatenate these four stiffnesses to form a vector $\frac{d\meshforce^{ijkl}}{d\nodes_{ijkl}\at{\idx}} \in \Real{12}$, which is the $\idx$-th column of the $12 \times 12$ vectorized element stiffness matrix.
The mesh energy $E_{mesh}$, force $\meshforce$, and stiffness $\frac{d\meshforce}{d\nodes}$ are the sums of the contributions of all elements $ijkl$.

\subsection{Tendon \Properties}
\label{app:fem-tendon}

The \emph{routing path} $i_1,\ldots,i_n$ is the set of node indices tendon $i$ is attached to; each of these nodes is a \emph{via point}. The \emph{routing} $\routing = [\node_{i_{1}},\ldots,\node_{i_{n}}]$ is the Cartesian location of each via point. The tendon length deformation $\tendondef$ is defined in terms of $\routing$ and rest length $\restlength_i$:
\begin{equation} \label{eq:tendon_length}
\begin{split}
    \tendondef_i &= \sum_{k=1}^{n-1}\|\routing_{k+1} - \routing_{k}\|_2 - \restlength_i
\end{split}
\end{equation}

Following \cite{Bern17iros}, the energy of tendon $i$ is a smooth polynomial in $\tendondef_i$ defined in terms of a small smoothing parameter $\epsilon$ (which we choose in practice to be zero) and tendon modulus $\kappa_t$ ($\kappa_{tendon}$ in the main text):
\begin{equation} 
E^i_{tendon} = 
\left\{
\begin{array}{ll}
0 & \text{if }\tendondef_i < -\epsilon\\
\frac{\kappa_t}{6\epsilon}\tendondef_i^3 + \frac{\kappa_t}{2}\tendondef_i^2 + \frac{\kappa_t\epsilon}{2}\tendondef_i + \frac{\kappa_t\epsilon^2}{6} & \text{if } \tendondef_i < \epsilon\\
\kappa_t\tendondef_i^2 + \frac{\kappa_t\epsilon^2}{3} & \text{otherwise}
\end{array}
\right.
\label{eq:tendon_energy}
\end{equation}
The tendon tension $\tension_i$ is the (scalar) derivative of tendon energy with respect to deformation:
\begin{equation}
\begin{split}
    \tension_i = \frac{dE^i_{tendon}}{d\tendondef_i}
\end{split}
\label{eq:tension}
\end{equation}
The change in deformation per movement of each via point, $\frac{d\tendondef_i}{d\routing}$, is equivalent to the sum of the unit vectors pointing from each via point to its neighbors (with trivial exceptions at the endpoints, where the contribution of a neighbor is 0):
\begin{equation} 
\begin{split}
    \frac{d\tendondef_i}{d\routing_k} &= \frac{\routing_{k-1} - \routing_{k}}{||\routing_{k-1} - \routing_{k}||_2}
    + \frac{\routing_{k+1} - \routing_{k}}{||\routing_{k+1} - \routing_{k}||_2}    
\end{split}
\label{eq:ddef_drouting}
\end{equation}
The force $\tendonforce^i = -\frac{dE^i_{tendon}}{d\routing} = -\frac{d\tendondef_i}{d\routing} \tension_i$.  
Tendon stiffness is the force Jacobian:
\begin{equation}
\begin{split}
    \frac{d\tendonforce^i}{d\routing} &= -\frac{d^2 E^i_{tendon} }{(d\routing)^2}\\
    &= \frac{d}{d\routing}(-\frac{d\tendondef_i}{d\routing} \tension_i)\\
    &= -\frac{d\tendondef_i}{d\routing} \left( \frac{d\tension_i}{d\routing} \right)\tran -\frac{d^2\tendondef_i}{(d\routing)^2} \tension_i\\
    &=-\frac{d^2 E^i_{tendon} }{d\tendondef_i^2} \frac{d\tendondef_i}{d\routing} 
    \left(\frac{d\tendondef_i}{d\routing}\right)\tran -\frac{d^2\tendondef}{(d\routing)^2} \tension_i
\end{split}
\label{eq:tendon_stiffness}
\end{equation}{}
This requires the second derivative of energy with respect to deformation $\frac{d^2 E^i_{tendon}}{d\tendondef_i^2}$, which is straightforward from \eqref{eq:tendon_energy}, as well as the gradient $\frac{d\tendondef_i}{d\routing}$ in \eqref{eq:ddef_drouting}. It also requires the tendon deformation Hessian $\frac{d^2\tendondef_i}{(d\routing)^2}$, which is the Jacobian of $\frac{d\tendondef_i}{d\routing}$~\eqref{eq:ddef_drouting}. Each element of \eqref{eq:ddef_drouting} is the sum of two unit vectors, so we first determine the $3\times 3$ Jacobian of a unit vector $\hat{\vq} = [q_1, q_2, q_3]\tran / \sqrt{q_1^2 + q_2^2 + q_3^2}$ with respect to its components:
\begin{equation} \label{eq:dUnitdX}
    \frac{d\hat{\vq}_{l}}{dq_m} = 
    \left\{
    \begin{array}{ll} 
        \frac{q_{l-1}^2 + q_{l-2}^2}{\sqrt{q_1^2 + q_2^2 + q_3^2}^3} & \text{if }l=m\\
        \\
        \frac{- q_{l} q_{m}}{\sqrt{q_1^2 + q_2^2 + q_3^2}^3} & \text{if }l \ne m
    \end{array}
    \right.
\end{equation}
The deformation Hessian
$\frac{d^2\tendondef_i}{(d\routing)^2}$ is sparse, with blocks on the main $3\times 3$ diagonal and those immediately above and below it. The main diagonal block $k$ is the sum of the Jacobians of the unit vectors towards the neighbors of $\routing_k$, the block above it is the negated Jacobian of the vector towards its predecessor, and the block below it is the negated Jacobian towards its successor. As above, there are trivial exceptions for endpoints. In other words, if we define $\vq^b_a$ as the unit vector from node $\routing_a$ to $\routing_b$ and $\hat{\vq}^b_a = \frac{\vq^b_a}{||\vq^b_a||}$ the structure of $\frac{d^2\tendondef_i}{(d\routing)^2}$ is as follows:
\begin{equation}
\begin{split}
&\frac{d^2\tendondef_i}{(d\routing)^2} = \\
&\begin{bmatrix}
\frac{d \hat{\vq}^{2}_{1}}{d \vq^{2}_{1}} & -\frac{d \hat{\vq}^{1}_{2}}{d \vq^{1}_{2}} & 0 & \dots &  0\\
-\frac{d \hat{\vq}^{2}_{1}}{d \vq^{2}_{1}} & \frac{d \hat{\vq}^{1}_{2}}{d \vq^{1}_{2}} + \frac{d \hat{\vq}^{3}_{2}}{d \vq^{3}_{2}} &  &  & 0\\
0 & -\frac{d \hat{\vq}^{3}_{2}}{d \vq^{3}_{2}} & \ddots & -\frac{d \hat{\vq}^{n-2}_{n-1}}{d \vq^{n-2}_{n-1}} & 0\\
\vdots & & &\frac{d \hat{\vq}^{n-2}_{n-1}}{d \vq^{n-2}_{n-1}} + \frac{d \hat{\vq}^{n}_{n-1}}{d \vq^{n}_{n-1}} & -\frac{d \hat{\vq}^{n-1}_{n}}{d \vq^{n-1}_{n}} \\
0 &  0 & 0 & -\frac{d \hat{\vq}^{n}_{n-1}}{d \vq^{n}_{n-1}} & \frac{d \hat{\vq}^{n-1}_{n}}{d \vq^{n-1}_{n}} \\
\end{bmatrix}
\end{split}
\end{equation}
This is used to compute the tendon stiffness as in \eqref{eq:tendon_stiffness}.

The tendon energy $E_{tendon}$, force $\tendonforce$, and stiffness $\frac{d\tendonforce}{d\nodes}$ are the sums of the contributions of all tendons.

\subsection{Pin \Properties}
\label{app:fem-pin}

Each pin $i$ is modeled as a spring with constant $\kappa_{pin}$, connecting a mesh node $\node_i$ belonging to the soft gripper, to a point $\vxx_i^{pin}$  belonging to the quadrotor base (for a given drone state $\MX$). 
The energy for each pin $i$ is as follows:
\begin{equation}
    E^i_{pin}(\node_i, \MX) = \kappa_{pin} \|\node_i - \vxx_i^{pin}\|_2
\end{equation}
The pin force $ \pinforce^i= -\frac{d E^i_{pin}}{d \node_i}=-\text{$\kappa$}_{pin}(\node_i - \vxx_i^{pin})$ and stiffness $\frac{d\pinforce^i}{d\node_i} = -\kappa_{pin} \eye$. 
The pin energy, force, and stiffness are the sums of the contributions of all pins.

\subsection{Gravity \Properties}
\label{app:fem-gravity}

We approximate the gripper mass as concentrated in the mesh nodes, and denote with $\nodemass_i$ the mass of node $i$.
The gravitational potential energy is determined by the mass and height of each node $i$:
\begin{equation}
    E^i_{gravity}(\nodes) = - \nodemass_i \; \vg\tran \node_i
\end{equation}
where $\vg \doteq [0,0,-9.81]\tran \text{m/s}^2$ is the gravity vector. Gravitational force $\gravityforce^i = \nodemass_i \vg$ and gravity has no stiffness.
The gravitational energy  and force are the sum of the contribution of all nodes.

\subsection{Actuator Jacobian}
\label{app:fem-jacobian}

Finally, our algorithm requires the actuator Jacobian $\frac{d\nodes}{d\restlengths}$. Our solution follows \cite{Bern17iros}.

First, we note that the quasi-static assumption defines a subspace on which the overall force $\Force$ is zero everywhere. Thus, all derivatives of $\Force$ are likewise $0$ on this subspace.
Changing the tendon rest lengths $\restlengths$ results in a change in tendon tensions $\tensions$ and node positions $\nodes$; the total derivative of $\Force$ with respect to $\restlengths$ yields partial derivatives in $\tensions$ and $\nodes$, which sum to zero:
\begin{equation}
\begin{split}
    \frac{d\Force}{d\restlengths} &= \frac{\delta \Force}{\delta \tensions} \frac{d\tensions}{d\restlengths} + \frac{\delta \Force}{\delta \nodes} \frac{d\nodes}{d\restlengths} = 0
\end{split}
\label{eq:linearsys}
\end{equation}

Besides $\frac{d\nodes}{d\restlengths}$, which is the quantity for which we are solving, all the remaining terms are known. $\frac{\delta \Force}{\delta \tensions}$ is exactly the aggregation of the matrices described in equation~\eqref{eq:ddef_drouting} for each tendon. $\frac{d\tensions}{d\restlengths}$ is  straightforward to compute from equations~\eqref{eq:tendon_length}, \eqref{eq:tendon_energy}, \eqref{eq:tension}. And $-\frac{\delta \Force}{\delta \nodes}$ is the sparse system Hessian $\frac{d^2E}{d\nodes^2}$:
\begin{equation}
\begin{split}
    \frac{d^2E}{d\nodes^2} = &-\frac{d\meshforce}{d\nodes} - \frac{d\tendonforce}{d\nodes} \\
    &-\frac{d\pinforce}{d\nodes} - \frac{d\gravityforce}{d\nodes} 
\end{split}
\end{equation}
Given these matrices, the resulting sparse linear system~\eqref{eq:linearsys} can be solved with any linear equation solver.


\section{Proof of Theorem~\ref{thm:controllerConvergence}}
\label{subsec:controller}

Here we prove that the quadrotor velocity and attitude, controlled as discussed in Section~\ref{sec:quadrotorBase}, converge to the desired values despite the presence of 
the soft gripper. The challenge lies in the fact that the soft load exerts a torque on the quadrotor center of mass 
which was not accounted for in the original geometric controller design.

\subsection{Outline}
We show that the attitude dynamics stabilize to a unique equilibrium, and the velocity error asymptotically approaches a limit proportional to the deviation of the attitude from equilibrium. Therefore, an unmodified geometric controller converges to a desired quadrotor attitude and velocity even in the presence of the  disturbance induced by the soft load.
The proof proceeds as follows:
\begin{itemize}
\item We analyze the tracking error and restate the theorem assumptions
 when restricting the quadrotor to the vertical plane. 
\item We show that the attitude stabilizes asymptotically to $\theta=\theta_{eq}$. In general, $\theta_{eq} = \theta_d$ iff. $\theta_d=0$.
\item We bound the attitude errors in terms of $d\theta$, the deviation of the attitude from $\theta_{eq}$; the previous section showed that $d\theta$  asymptotically approaches zero.
\item We bound the asymptotic magnitude of the total velocity error proportional to the horizontal velocity error and $|d\theta|$. 
\item We show that the horizontal velocity error asymptotically approaches zero. In light of the previous section, this further implies that the total velocity error vanishes asymptotically; however, we show explicitly that the vertical velocity error vanishes as well in the next section.
\item Finally, we show that the vertical velocity error also asymptotically approaches zero.
\end{itemize}

\subsection{Tracking Errors and Assumptions in the Plane}
\label{subsubsec:quad_setup}
As is common in aerial manipulation (see, \eg Thomas \setal~\cite{Thomas14bioinspiration}), we consider a case in which the quadrotor is confined to the vertical plane, with the quadrotor velocity and attitude denoted as $(v_x, v_z , \theta)$ (Fig.~\ref{fig:CM}).
We denote the desired velocity and attitude as $(v_{xd}, v_{zd}, \theta_{d})$ where $\theta_d$ is chosen such that velocity converges to the desired velocity (as described in \eqref{eq:theta_d} below). This allows us to express all control quantities defined on the rotation manifold $\SOtwo$ in terms of a single angle.
In particular, the rotation errors in (\ref{eq:trackingErrors}) simplify to elementary trigonometric functions:
\begin{equation}
\begin{split}
    \Psi &= 1 - \cos(\theta - \theta_d) \\
    e_r &= \frac{d\Psi}{d\theta} = \sin(\theta - \theta_d) \\
    e_{\Omega} &= \Dot{\theta} -\Dot{\theta}_d \\
\end{split}
\label{eq:attitude_errors}
\end{equation}
Similarly, the velocity errors become:
\begin{equation}
\begin{split}
    \ve_v = [e_{v_x}, e_{v_z}]\tran = [v_{x}, v_{z}]\tran - [v_{xd}, v_{zd}]\tran
\end{split}
\label{eq:ev}
\end{equation}
And the quadrotor body frame $\MR$ is defined solely in terms of the angle $\theta$:
\begin{equation}
\begin{split}
      R &= [\rotcol_x, \rotcol_z]\\
      &= 
      \begin{bmatrix} \cos(\theta) & -\sin(\theta) \\
                      \sin(\theta) & \cos(\theta) \\ \end{bmatrix}
\end{split}
\label{eq:quad_frame_2d}
\end{equation}

We define the load center of mass position relative to the quadrotor center of mass $x_{\load}, z_{\load}$, mass $m_{\load}$ and distance from the attachment point to the load center of mass when the quadrotor is vertical $\lenCOM$ (Fig.~\ref{fig:CM}).
\begin{figure}
    \centering
    \includegraphics[width=.45\textwidth]{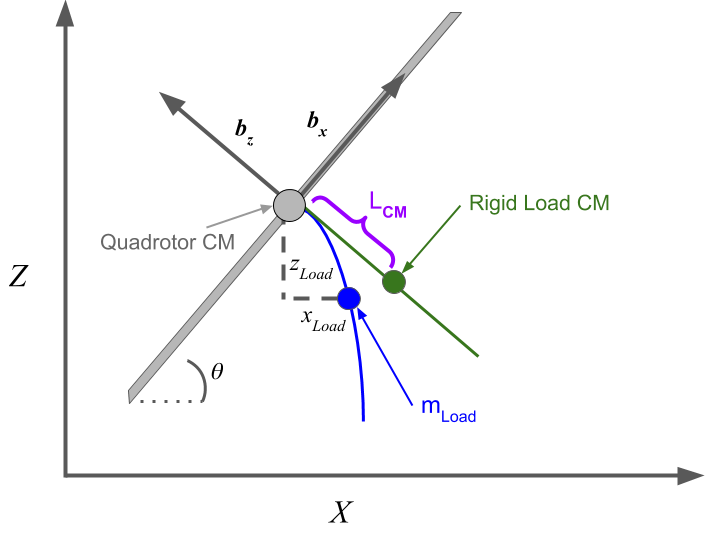}
    \caption{Quadrotor confined to the vertical plane. The figure shows the quadrotor body frame $\MR = [\rotcol_x, \rotcol_z]$, angle $\theta$, soft load center of mass relative to the quadrotor $x_{Load}, z_{Load}$, and the rigid load centers of mass (also compare to Fig.~\ref{fig:geoControl}(b)).\vspace{-5mm}\label{fig:CM}}
\end{figure}

{
Without aerodynamic drag, the load does not deform and there is no uncompensated load torque. Drag (assumed to act on the quadrotor center of mass) results in a deformation of the load and an uncompensated torque on the quadrotor (Fig. \ref{fig:drag_deformation}). This can act either to increase or decrease $\theta$, depending on whether the $x$ components of velocity and thrust are aligned. However, the state in which velocity and thrust point in opposite directions is inherently transient (because both drag and thrust act against velocity, thus quickly reducing velocity and drag); similarly for the case in which drag forces exceed thrust forces. Our assumptions on the load torque in~\eqref{eq:quad_assumptions_1} focus on the limiting case in which velocity and acceleration are aligned and the quadrotor has achieved static equilibrium (zero acceleration), but apply to any scenario in which the drag force is opposite and smaller than the thrust force (or does not exist at all).
}

\begin{figure}[t]
    \begin{center}
    \begin{minipage}{\textwidth}
    \begin{tabular}{ccc}%
    \myhspace \hspace{-5mm}
            \begin{minipage}{\mpw}%
            \centering%
            \includegraphics[trim=0mm 0mm 0mm 0mm, clip,height=3cm]{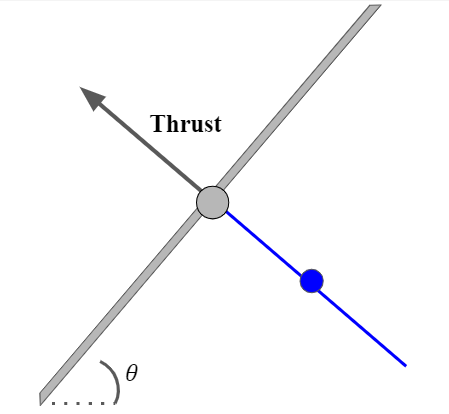} \\
            \hspace{0cm}(a) 
            \end{minipage}
        & \hspace{-20mm}
            \begin{minipage}{\mpw}%
            \centering%
            \includegraphics[trim=0mm 0mm 0mm 0mm, clip,height=3cm]{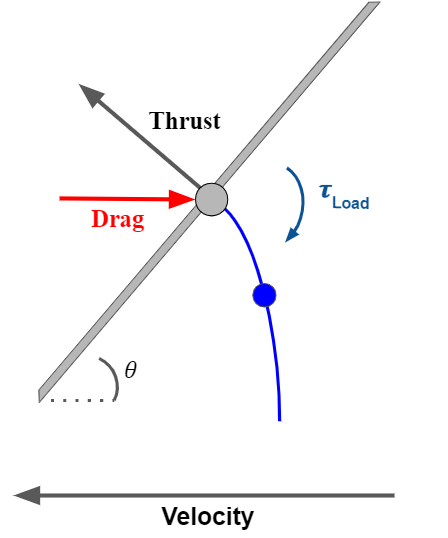} \\
            \hspace{0cm}(b) 
            \end{minipage}
        & \hspace{-20mm}
            \begin{minipage}{\mpw}%
            \centering%
            \includegraphics[trim=0mm 0mm 0mm 0mm, clip,height=3cm]{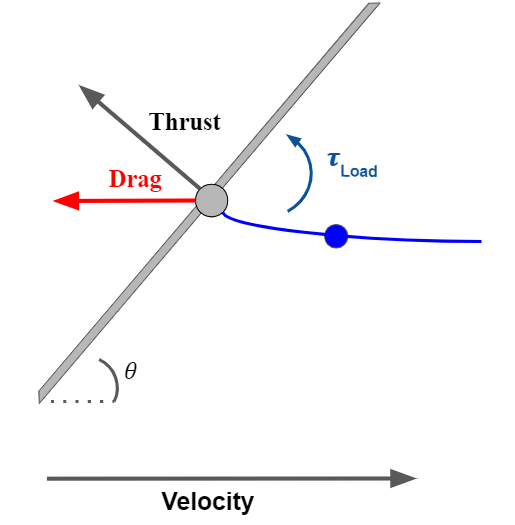} \\
            \hspace{0cm}(c) 
            \end{minipage}
        \end{tabular}
    \end{minipage}
    \begin{minipage}{\textwidth}
    \end{minipage}
    \vspace{-2mm} 
    \caption{
    {(a) With no aerodynamic force and no torque, the load does not deform and exerts no torque.
    (b) With aerodynamic forces opposing thrust forces, the load deforms and exerts a torque which decreases the quadrotor angle.
    (c) With aerodynamic forces acting with thrust forces, the load exerts a torque which increases the quadrotor angle.}
    \label{fig:drag_deformation}\vspace{-7mm} }
    \end{center}
\end{figure}

We denote the magnitude of the load's first moment of mass with $\loadml = m_{\load} \lenCOM$. We further define $g \doteq \|\vg\|$, which is the norm of the gravity vector.
With this notation, the assumptions in~\eqref{eq:quad_assumptions_1} reduce to: 
\bea
    -\pi/2 < \theta <\pi/2  \;\; &\text{and}& \;\; -\pi/2 < \theta_d <\pi/2 \label{eq:a1}  \\
    |\tau_{\load}| &\leq& \loadml  g \; |\sin(\theta)| \label{eq:a2} \\
    \sign (\tau_{\load}) &=& -\sign(\theta) \label{eq:a3} 
\eea
Intuitively,~\eqref{eq:a1} requires the the drone is not upside-down, while~\eqref{eq:a2}-\eqref{eq:a3} 
require that (i) when the quadrotor is tilted, the load deforms under gravity such that its center of mass is lower and closer to the vertical than that of a corresponding rigid load (Fig.~\ref{fig:geoControl} and Fig.~\ref{fig:CM}), 
and (ii) at rest and with the quadrotor level, the load center of mass is directly below the quadrotor center of mass.
This implies that the torque $\tau_{\load}$ (exerted by the soft load) 
is always of the opposite sign as $\theta$ and is upper-bounded by the torque exerted by a rigid load. 
We remark that these assumptions are satisfied by a symmetric soft load like that considered in this paper.

We further define $E_{\load}$ as the gravitational potential energy of the load in the non-rotating frame fixed to the quadrotor center of mass. 
Under our assumptions this is symmetric about $\theta = 0$ and upper-bounded by the energy of a rigid load:
\begin{equation}
\begin{split}
    E_{\load} &=  m_{\load} \; z_{\load} \; g\\
    &\leq  - \loadml \; \cos(\theta) \; g \\
\end{split}
\label{eq:E_load}
\end{equation}
By conservation of energy, the gradient of gravitational potential $E_{\load}$ with respect to $\theta$ corresponds to the gravitational torque $\tau_{\load}$:
\begin{equation}
    \frac{d E_{\load}}{d \theta} = - \tau_{\load}
\end{equation}


\subsection{Attitude Stability}
\label{subsubsec:att_stable}
Given moment of inertia $J$, the attitude dynamics (restricted to the vertical plane) are: 
\begin{equation}
\begin{split}
    J \Ddot{\theta} = \tau + \tau_{\load}
\end{split}
\label{eq:attitude_dynamics_1}
\end{equation}
We choose the control $\tau$ as in~\eqref{eq:attitude_errors}:
\begin{equation}
\begin{split}
    \tau = -k_r e_r - k_{\Omega}e_{\Omega} + J\Ddot{\theta}_d
\end{split}
\label{eq:attitude_controller}
\end{equation}
where $k_r$, $k_{\Omega}$ are the control gains.
Substituting the controller~\eqref{eq:attitude_controller} back into the dynamics~\eqref{eq:attitude_dynamics_1}, the closed-loop attitude dynamics become:
\begin{equation}
\begin{split}
     J \Ddot{\theta} &= -k_r e_r - k_{\Omega}e_{\Omega} + J\Ddot{\theta}_d + \tau_{\load} \\
\label{eq:attitude_dynamics_2}
\end{split}
\end{equation}

We show that, under assumptions~\eqref{eq:a1}-\eqref{eq:a3}, the closed-loop system~\eqref{eq:attitude_dynamics_2} is stable. We define a Lyapunov function $V_{\theta}$ and show that this is negative semi-definite so the attitude asymptotically approaches the { equilibrium angle $\theta= \theta_{eq}$}.
A Lyapunov function $V_\theta$ can be defined by analogy to the total energy of a double pendulum, where the upper pendulum stabilizes to ${\theta}_d$ rather than the vertical, the lower pendulum is non-rigid and in quasi-static equilibrium and there is angular velocity damping proportional to $e_{\Omega}$:
\begin{equation}
\begin{split}
     V_\theta &= \frac{J}{2} e_{\Omega}^2 +  k_r \Psi + E_{\load} \\
\end{split}
\label{eq:V_theta}
\end{equation}
We compute the gradient of $V_\theta$ using \eqref{eq:attitude_dynamics_2}, \eqref{eq:attitude_errors} and \eqref{eq:E_load}:
\begin{equation}
\begin{split}
    \Dot{V_\theta} &= e_{\Omega} (J\Ddot{\theta} - J\Ddot{\theta}_d) + k_r e_r e_{\Omega} - \tau_{\load} e_{\Omega} \\
    &= e_{\Omega} (-k_r e_r - k_{\Omega}e_{\Omega} + \tau_{\load}) + k_r e_r e_{\Omega} - \tau_{\load} e_{\Omega} \\
    &= -k_{\Omega}e_{\Omega}^2 \leq 0
\end{split}
\end{equation}

The Lyapunov function $V_\theta$ is negative semi-definite, so the system will converge to the largest invariant set $\{ e_{\Omega} = 0, \Dot{e}_{\Omega} = 0 \}$; we substitute these values into \eqref{eq:attitude_dynamics_2} to find the equilibrium angle $\theta_{eq}$:
\begin{equation}
\begin{split}
    J \Ddot{\theta}  - J\Ddot{\theta}_d  +   k_{\Omega}e_{\Omega}&= -k_r e_r + \tau_{\load} \\
    J \Dot{e}_{\Omega} +  k_{\Omega}e_{\Omega}&= -k_r e_r + \tau_{\load} \\
    0 &= -k_r e_r(\theta_{eq}) + \tau_{\load}(\theta_{eq}) \\
    \tau_{\load}(\theta_{eq}) &= k_r \sin(\theta_{eq} - \theta_d) \\
\end{split}
\label{eq:equilibrium_angle_1}
\end{equation}

In general, the stable angle $\theta_{eq}$ is not equal to the desired angle $\theta_d$. We establish the relationship between these two angles below.
Because $\tau_{\load}$ acts to decrease the magnitude of $\theta$ \eqref{eq:a3}, it is clear that $|\theta_{eq}| \leq |\theta_d|$. This implies that $\sign \sin(\theta_{eq} - \theta_d) = -\sign \theta_d$. We denote $\sign \theta_d$ as $\sign_{\theta_d}$ and show that this is equal to $\sign \theta_{eq}$ using \eqref{eq:a3}, \eqref{eq:equilibrium_angle_1}: 
\begin{equation}
\begin{split}
      \sign (\tau_{\load}) &= \sign k_r \sin(\theta_{eq} - \theta_d) \\
      -\sign \theta_{eq} &= \sign \sin(\theta_{eq} - \theta_d) \\
     \sign \theta_{eq} &= \sign_{\theta_d} \\
\end{split}
\label{eq:equilibrium_angle_sign} 
\end{equation}
We use \eqref{eq:a2}, \eqref{eq:equilibrium_angle_sign} to bound \eqref{eq:equilibrium_angle_1}:
\begin{equation}
\begin{split}
    \loadml \,g\, \ |\sin(\theta_{eq})| &\geq k_r |\sin(\theta_{eq} - \theta_d)| \\
    \loadml \,g\, |\sin(\theta_{eq})| &\geq -k_r \sin(\theta_{eq} - \theta_d) \sign_{\theta_d}
\end{split}
\label{eq:equilibrium_angle_2}
\end{equation}
Which we expand using the trigonometric identity $\sin(a-b) = \sin(a) \cos(b) - \cos(a) \sin(b)$: 
\begin{equation}
\begin{split}
    \loadml \,g\, |\sin(\theta_{eq})| &\geq  -k_r (\sin(\theta_{eq}) \cos(\theta_d) \\
    & \quad - \cos(\theta_{eq}) \sin(\theta_d)) \sign_{\theta_d}  \\
    \loadml \,g\, |\sin(\theta_{eq})| &\geq  -k_r (|\sin(\theta_{eq})| \cos(\theta_d) \\
    & \quad - \cos(\theta_{eq}) |\sin(\theta_d)|)   \\
\end{split}
\label{eq:equilibrium_angle_3}
\end{equation}
We solve \eqref{eq:equilibrium_angle_3} for a bound on the magnitude of $\theta_{eq}$ as a function of $\theta_d$:
\begin{equation}
\begin{split}
    \loadml \,g &\geq  -k_r (|\sin(\theta_{eq})| \cos(\theta_d) \\
    & \quad - \cos(\theta_{eq}) |\sin(\theta_d)|)/|\sin(\theta_{eq})|   \\
    \loadml \,g\  &\geq -k_r \left( \cos(\theta_d)  - \frac{|\sin(\theta_d)|}{|\tan(\theta_{eq})|} \right) \\
    |\tan(\theta_{eq})|  &\geq \frac{|\sin (\theta_d)|}{\frac{\loadml \,g}{k_r} + \cos(\theta_d)}  \\  
    |\theta_{eq}| &\geq \left|\tan^{-1} \left(\frac{\sin (\theta_d)}{\frac{\loadml \,g}{k_r} + \cos(\theta_d)} \right) \right|
\end{split}
\label{eq:equilibrium_angle}
\end{equation}
Thus, we have $|\theta_d| \ \geq |\theta_{eq}| \ \geq  \left|\tan^{-1} \left(\frac{\sin (\theta_d)}{(\loadml \,g\,)/k_r + \cos(\theta_d)} \right) \right|$. When $\theta_d$ is zero, these bounds are equal and $\theta_{eq} = \theta_d = 0$; otherwise $\theta_{eq} \neq \theta_d$.

\subsection{Bounding the Rotation Error}
\label{subsubsec:bound_rot_err}
In the previous section, we have shown convergence to $\theta_{eq}$ rather than $\theta_d$. This means that there exists some equilibrium rotation error $e_r(\theta_{eq})$. Further, it is convenient to express $e_r(\theta)$ in general in terms of $d\theta = \theta - \theta_{eq}$, rather than as $\theta - \theta_d$. We provide upper bounds for both these terms here. 
To simplify notation, in the following we denote $\sin(d\theta)$ by $s_{d\theta}$.

We can bound elementary trigonometric functions of $\theta_{eq}$ using \eqref{eq:equilibrium_angle} and the definition of the tangent:
\bea
|\sin(\theta_{eq})| &\geq&  \left| \frac{\sin(\theta_d)}{\sqrt{\left( \frac{\loadml \,g\,}{k_r} \right)^2 + 2 \frac{\loadml \,g\,}{k_r} \cos (\theta_d) + 1}}  \right| \label{eq:s_eq} \\
|\cos(\theta_{eq})| &\leq& \left| \frac{\frac{\loadml \,g\,}{k_r} + \cos(\theta_d)}{\sqrt{\left( \frac{\loadml \,g\,}{k_r} \right)^2 + 2 \frac{\loadml \,g\,}{k_r} \cos (\theta_d) + 1}} \right| \label{eq:c_eq}
\eea
We define a constant $\eta = \frac{\frac{\loadml \,g\,}{k_r} } {\sqrt{\left( \frac{\loadml \,g\,}{k_r} \right)^2 + 1}}$. From \eqref{eq:s_eq},~\eqref{eq:c_eq} and \eqref{eq:a1}, we show that the rotation error associated with the equilibrium angle, $e_r(\theta_{eq})$, can be bounded as a function of $\sin(\theta_d)$:
\begin{equation}
\begin{split}
    |e_r(\theta_{eq})| &= |\sin(\theta_{eq} - \theta_{d})| \\
    &= |\sin(\theta_{eq})\cos(\theta_d) - \cos(\theta_{eq})\sin(\theta_d)| \\
    &\leq {\left| \frac{\sin(\theta_d) \cos(\theta_d) - [\frac{\loadml \,g\,}{k_r} + \cos(\theta_d)]\sin(\theta_d)} {\sqrt{\left( \frac{\loadml \,g\,}{k_r} \right)^2 + 2 \frac{\loadml \,g\,}{k_r} \cos (\theta_d) + 1}} \right|} \; \text{\footnotemark}\\
    &\leq \frac{\frac{\loadml \,g\,}{k_r} } {\sqrt{\left( \frac{\loadml \,g\,}{k_r} \right)^2 + 2 \frac{\loadml \,g\,}{k_r} \cos (\theta_d) + 1}} |\sin (\theta_d)| \\
    &\leq \eta |\sin (\theta_d)|
\end{split}
\label{eq:er_eq}
\end{equation}
\footnotetext{[Footnote to Eq.~\ref{eq:er_eq}]: This is non-trivial. Assume without loss of generality that $\theta_d , \theta_{eq}$ are positive; otherwise we can multiply by $\sign_{\theta_d}$ as we do elsewhere. This implies that $\sin(\theta_d), \sin(\theta_{eq})$ are positive;  $\cos(\theta_d), \cos(\theta_{eq})$ are positive from \eqref{eq:a1}. Further, from \eqref{eq:equilibrium_angle_sign} $\sin(\theta_{eq} - \theta_{d})$ is negative if $\theta_d , \theta_{eq}$ are positive. We have $\sin(\theta_{eq} - \theta_{d}) = \sin(\theta_{eq})\cos(\theta_d) - \cos(\theta_{eq})\sin(\theta_d)$, so  $\cos(\theta_{eq})\sin(\theta_d)$ must be larger than $\sin(\theta_{eq})\cos(\theta_d)$. Therefore, to maximize the magnitude of $\sin(\theta_{eq} - \theta_{d})$ we upper-bound $\cos(\theta_{eq})$ and lower-bound $\sin(\theta_{eq})$. These are the bounds provided by \eqref{eq:s_eq},~\eqref{eq:c_eq}.}
And from \eqref{eq:er_eq}, $e_r(\theta)$ in general can be bounded as:
\begin{equation}
\begin{split}
    |e_r(\theta)| &= |\sin(\theta_{eq} - \theta_d + d\theta)| \\
    &\leq |\sin(\theta_{eq} - \theta_d)| + |s_{d\theta}| \\
    &\leq  \eta |\sin (\theta_d)| + |s_{d\theta}|
\end{split}
\label{eq:er_bound}
\end{equation}
These allow us to discuss the evolution of rotation error $e_r$ as $d\theta$ asymptotically approaches 0.

\begin{figure}
    \centering
    \includegraphics[width=.4\textwidth]{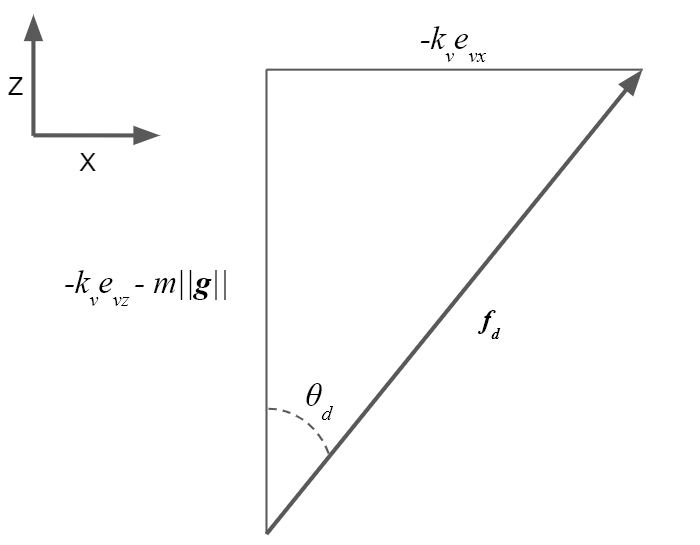}
    \caption{The desired force $\fdes$ and desired angle $\theta_d$.}
    \label{fig:fdes}
\end{figure}

\subsection{Bounding the Total Velocity Error}
\label{subsubsec:vel_thresh}
The velocity error dynamics in the vertical plane can be written as:

\begin{equation}
\begin{split}
    m\Dot{\ve_v} &= m \vg  + f \rotcol_z
\end{split}
\label{eq:velocity_dynamics_1}
\end{equation}
We choose the desired thrust force $\fdes$ and actual thrust force $f$ as in the geometric controller equations~\eqref{eq:geoController}, which when restricted to the vertical plane and under the assumptions of the theorem becomes:
\bea
    \fdes &=& -k_v \ve_v - m \vg \label{eq:f_des} \\
    f &=& \fdes\cdot \rotcol_z \label{eq:f}
\eea
We choose the desired angle $\theta_d$ based on \eqref{eq:f_des} to align $\rotcol_z$ with the desired force $\fdes$ (Fig.~\ref{fig:fdes}).
\begin{equation}
\begin{split}
    \theta_d = \sin^{-1} \left(\frac{-k_v e_{v_z}}{||\fdes||}\right)
\end{split}
\label{eq:theta_d}
\end{equation}
We define rotation errors and choose torques as in the previous section such that $\theta$ asymptotically approaches $\theta_{eq}$.  

Similarly to Lee~\setal\cite{Lee10arxiv}, we rewrite the closed-loop dynamics in terms of attitude error $e_r$. First we add and subtract $\frac{f}{\cos(\theta - \theta_d)} \frac{\fdes}{||\fdes||}$:
\begin{equation}
\begin{split}
    m\Dot{\ve_v} &= m \vg + \frac{f}{\cos(\theta - \theta_d)} \frac{\fdes}{||\fdes||} \\
    & \quad + \frac{f}{\cos(\theta - \theta_d)} \left(\cos(\theta - \theta_d)\rotcol_z - \frac{\fdes}{||\fdes||} \right) \\
    &= m \vg + \frac{f}{\cos(\theta - \theta_d)} \frac{\fdes}{||\fdes||}  + \frac{f}{\cos(\theta - \theta_d)} \vw
\end{split}
\label{eq:velocity_dynamics_2}
\end{equation}
Where $\vw$ is defined as:
\begin{equation}
    \vw = \cos(\theta - \theta_d)\rotcol_z - \frac{\fdes}{||\fdes||} 
\label{eq:w}
\end{equation}
$\frac{\fdes}{||\fdes||}$ is the desired $\rotcol_z$ orientation, which from \eqref{eq:theta_d}  is $[-\sin(\theta_d), \cos(\theta_d)]\tran$. From \eqref{eq:quad_frame_2d} $\rotcol_z = [-\sin(\theta), \cos(\theta)]\tran$. We substitute these values into \eqref{eq:w}.
\begin{equation}
\begin{split}
    \vw &= \cos(\theta - \theta_d) \begin{bmatrix} -\sin(\theta) \\ \cos(\theta) \end{bmatrix} - \begin{bmatrix} -\sin(\theta_d) \\ \cos(\theta_d) \end{bmatrix} \\
\end{split}
\label{eq:w1}
\end{equation}
Using common trigonometric identities, we show that $\vw$~\eqref{eq:w1} is proportional to the rotation error $e_r$ and aligned with the quadrotor axis $\rotcol_x = [\cos(\theta), \sin(\theta)]\tran$~\eqref{eq:quad_frame_2d}.
\begin{equation}
\begin{split}
    \vw &= (\sin(\theta)\sin(\theta_d) + \cos(\theta)\cos(\theta_d)) \begin{bmatrix} -\sin(\theta) \\ \cos(\theta) \end{bmatrix} \\
    &\quad- \begin{bmatrix} -\sin(\theta_d) \\ \cos(\theta_d) \end{bmatrix} \\
    &= \begin{bmatrix} \sin(\theta_d) (1 - \sin^2(\theta)) - \sin(\theta) \cos(\theta_d) \cos(\theta) \\ -\cos(\theta_d) (1 - \cos^2(\theta))  + \cos(\theta) \sin(\theta_d) \cos(\theta)\end{bmatrix} \\
    &= \begin{bmatrix} \sin(\theta_d) \cos^2(\theta) - \sin(\theta) \cos(\theta_d) \cos(\theta) \\ -\cos(\theta_d) \sin^2(\theta)  + \cos(\theta) \sin(\theta_d) \cos(\theta)\end{bmatrix}  \\
    &= (\sin(\theta_d)\cos(\theta) - \cos(\theta_d)\sin(\theta)) \begin{bmatrix} \cos(\theta) \\ \sin(\theta) \end{bmatrix} \\
    &= \sin(\theta_d - \theta) \begin{bmatrix} \cos(\theta) \\ \sin(\theta) \end{bmatrix} \\
    &= - e_r \rotcol_x
\end{split}
\label{eq:w2}
\end{equation}
From \eqref{eq:f} $||\fdes|| = \frac{f}{\cos(\theta - \theta_d)}$. This, with \eqref{eq:w2} and \eqref{eq:f_des}, allows us to reduce the closed loop dynamics \eqref{eq:velocity_dynamics_2} to the desired correction term proportional to the velocity error, as well as a disturbance term along the quadrotor $\rotcol_x$ axis proportional to the magnitude of the desired force $\fdes$ and the rotation error $e_r$:
\begin{equation}
\begin{split}
    m\Dot{\ve_v} &= m \vg + ||\fdes|| \frac{\fdes}{||\fdes||}  + ||\fdes|| \vw \\
    &= m \vg + \fdes - ||\fdes|| e_r \rotcol_x \\
    &= -k_v \ve_v - ||\fdes|| e_r \rotcol_x
\end{split}
\label{eq:velocity_dynamics}
\end{equation}

We show that it is possible to define a bound defined by the horizontal velocity error $e_{v_x}$ and the deviation from equilibrium attitude $d\theta$, above which the velocity error decreases. In order to do so we define a Lyapunov candidate $V_v$ and show that this is negative definite when $||\ve_v||$ exceeds some threshold.
Consider the Lyapunov candidate $V_v$, which if negative definite would demonstrate velocity convergence:
\bea
    V_v &= \frac{1}{2} m ||\ve_v||^2 \label{eq:Vv} \\
    \Dot{V_v} &=  \ve_v \cdot m\Dot{\ve_v} \label{eq:Vv_dot_1} \\
\eea
We can upper-bound $\Dot{V_v}$~\eqref{eq:Vv_dot_1} using \eqref{eq:velocity_dynamics}:
\begin{equation}
\begin{split}
\Dot{V_v} &=  \ve_v \cdot (-k_v \ve_v - ||\fdes|| e_r \rotcol_x ) \\
&\leq ||\ve_v||(-k_v ||\ve_v|| + ||\fdes|| \,|e_r|) 
\end{split} 
\label{eq:Vv_dot_2}
\end{equation}
We further upper-bound $\Dot{V_v}$ by considering the term $||\fdes|| \, |e_r|$. As in Appendix~\ref{subsubsec:bound_rot_err}, we define $d\theta = \theta - \theta_{eq}$ and denote $\sin(d\theta)$ by $s_{d\theta}$. First, from \eqref{eq:theta_d} the horizontal component of the desired force is equal to $-\sin(\theta_d)||\fdes||$:
\begin{equation}
    |\sin(\theta_d)| \, ||\fdes|| = k_v |e_{v_x}| 
\label{eq:horiz_force}
\end{equation}
Equations \eqref{eq:horiz_force} and \eqref{eq:er_bound} allow us to bound $||\fdes|| \,|e_r| $ as jointly affine in $e_{v_x}$, $s_{d\theta}$:
\begin{equation}
\begin{split}
    ||\fdes|| \,|e_r|  &\leq ||\fdes|| (\eta |\sin (\theta_d|) + |s_{d\theta}|)\\
    &\leq k_v \eta |e_{v_x}| + ||\fdes|| \ |s_{d\theta}|
\end{split}
\label{eq:fdes_er_bound}
\end{equation}
Further, using \eqref{eq:f_des} we can bound $||\fdes||$ in terms of its components:
\begin{equation}
\begin{split}
    ||\fdes|| \leq k_v ||\ve_v|| + m \,g\, 
\end{split}
\label{eq:fdes_bound}
\end{equation}
Finally, we can bound $\Dot{V_v}$~\eqref{eq:Vv_dot_2} in terms of $e_{v_x}, s_{d\theta}$ by using \eqref{eq:fdes_er_bound},~\eqref{eq:fdes_bound}:
\begin{equation}
\begin{split}
    \Dot{V_v} &\leq ||\ve_v||(-k_v ||\ve_v|| + ||\fdes|| \,|e_r|)   \\
    &\leq  ||\ve_v||(-k_v ||\ve_v|| + k_v \eta |e_{v_x}| + ||\fdes|| \ |s_{d\theta}|)  \\
    &\leq  ||\ve_v||(-k_v ||\ve_v|| + k_v \eta |e_{v_x}| + (k_v ||\ve_v|| + m \,g\, ) \ |s_{d\theta}|)  \\
    &\leq ||\ve_v||(-k_v ||\ve_v||(1 -|s_{d\theta}|)  + k_v |e_{v_x}| \eta  +  m \,g\, \ |s_{d\theta}|) \\
\end{split}
\label{eq:Vv_dot_3}
\end{equation}
When $\ve_v$ exceeds some threshold $e_v^{stable}$, $\Dot{V_v}$ is negative definite and $||\ve_v||$ decreases monotonically. This occurs when the right-hand-side of \eqref{eq:Vv_dot_3} is negative:
\begin{equation}
\begin{split}
    0 &\geq ||\ve_v||(-k_v ||\ve_v||(1 -|s_{d\theta}|)  + k_v |e_{v_x}| \eta  +  m \,g\, \ |s_{d\theta}|) \\
    0 &\geq -k_v ||\ve_v||(1 -|s_{d\theta}|)  + k_v |e_{v_x}| \eta  +  m \,g\, \ |s_{d\theta}|\\
    ||\mathbf{e_v}|| &\geq \frac{|e_{v_x}| \eta + \frac{m \,g\,}{k_v} |s_{d\theta}|}{1 -|s_{d\theta}|} \\
    &\doteq e_v^{stable}
\end{split}
\label{eq:velocity_bound}
\end{equation}
Equation \eqref{eq:velocity_bound} bounds the norm of total velocity error $||\ve_v||$, but $e_{v_x}$ appears in the bound so velocity stability has not yet been shown. In the next section, we use the results above to show asymptotic convergence to the desired horizontal velocity as $\theta$ approaches $\theta_{eq}$.   

\subsection{Bounding Horizontal Velocity}
\label{subsubsec:horiz_bound}
In the previous section we defined a bound on $||\ve_v||$ affine in $|e_{v_x}|, |s_{d\theta}|$ above which velocity error decreases. However, clearly $|e_{v_x}| \leq ||\ve_v||$, so $|e_{v_x}| \geq e_v^{stable}$ implies $||\ve_{v}|| \geq e_v^{stable}$. Therefore, we can substitute $|e_{v_x}|$ for $||\ve_v||$ in (\ref{eq:velocity_bound}) to define a threshold $e_{v_x}^{stable}$ proportional only to $d\theta$, above which the magnitude of horizontal velocity error $|e_{v_x}|$ decreases monotonically:
\begin{equation}
\begin{split}
    |e_{v_x}| &> \frac{|e_{v_x}| \eta + \frac{m \,g\,}{k_v} |s_{d\theta}|}{1 - |s_{d\theta}|} \\
    |e_{v_x}|(1 - |s_{d\theta}| - \eta) &>\frac{m \,g\,}{k_v} |s_{d\theta}| \\
    |e_{v_x}| &> \frac{\frac{m \,g\,}{k_v} |s_{d\theta}|}{(1 - |s_{d\theta}| - \eta)} \\
    &\doteq e_{v_x}^{stable}
\end{split}
\label{eq:horizontal_bound}
\end{equation}
$|e_{v_x}|$ decreases monotonically to $e_{v_x}^{stable}$; this threshold is linear in $|s_{d\theta}|$, so as $d\theta$ asymptotically approaches zero $e_{v_x}$ does likewise.
Note that \eqref{eq:theta_d} implies that $\theta_d$ approaches zero with $e_{v_x}$, and that from \eqref{eq:equilibrium_angle} $\theta_d=0$ implies $\theta_{eq}=\theta_d$. Therefore, horizontal velocity convergence also implies that rotation error $e_r$ vanishes asymptotically. 

Equation (\ref{eq:velocity_bound}) provides a bound affine in $|e_{v_x}|, |s_{d\theta}|$ above which velocity error decreases monotonically. We have now shown that both of these terms asymptotically approach zero; this is sufficient to establish that velocity error likewise vanishes over time. However, we also show explicitly below that vertical velocity error $e_{v_z}$ vanishes. 


\subsection{Bounding Vertical Velocity}
\label{subsubsec:vert_bound}
Given the previous results, we show that there exists a bound on $e_{v_z}$ affine in $|e_{v_x}|, |s_{d\theta}|$ above which it decreases monotonically. Based on (\ref{eq:velocity_dynamics_2}) and the definition $\rotcol_x = [\cos(\theta), \sin(\theta)]\tran$, the vertical velocity dynamics are:
\begin{equation}
\begin{split}
     m\Dot{e_{v_z}} &= -k_v e_{v_z} - ||\fdes|| e_r \sin(\theta) \\
\end{split}
\label{eq:vert_dynamics}
\end{equation}

Consider the Lyapunov candidate $V_{v_z}$:
\begin{equation}
\begin{split}
    V_{v_z} &= \frac{1}{2} m e_{v_z}^2 \\
    \Dot{V_{v_z}} &= e_{v_z} (m\Dot{e_{v_z}})
\end{split}
\label{eq:Vvz}
\end{equation}
As in (\ref{eq:Vv_dot_3}), we can use \eqref{eq:fdes_er_bound},~\eqref{eq:fdes_bound},~\eqref{eq:vert_dynamics} to bound $\Dot{V_{v_z}}$~\eqref{eq:Vvz}:
\begin{equation}
\begin{split}
    \Dot{V_{v_z}} &\leq |e_{v_z}| (-k_v |e_{v_z}| \\ 
     & \quad+ k_v |e_{v_x}| \eta + (k_v ||\ve_v|| + m \,g\,)|s_{d\theta}|) \\
\end{split}
\label{eq:vert_velocity_bound_1}
\end{equation}
From \eqref{eq:ev} we have $||\ve_v|| \leq |e_{v_x}| + |e_{v_z}|$, so similarly to (\ref{eq:velocity_bound}) we set the right-hand-side of (\ref{eq:vert_velocity_bound_1}) to zero in order to find a bound above which $e_{v_z}$ decreases monotonically:
\begin{equation}
\begin{split}
    0 & \geq -k_v |e_{v_z}| \\ 
    & \quad+ k_v |e_{v_x}| \eta + (k_v ||\ve_v|| + m \,g\,)|s_{d\theta}| \\
    k_v |e_{v_z}| &\geq k_v |e_{v_x}| \eta + (k_v ||\ve_v|| + m \,g\,)|s_{d\theta}| \\
    k_v |e_{v_z}| &\geq k_v |e_{v_x}| \eta + (k_v |e_{v_x}| +  k_v |e_{v_z}| + m \,g\,)|s_{d\theta}| \\
    k_v |e_{v_z}| (1 - |s_{d\theta}|) &\geq (\eta + |s_{d\theta}|)k_v |e_{v_x}| + m \,g\, \ |s_{d\theta}| \\
    |e_{v_z}| &\geq \frac{(\eta + |s_{d\theta}|) |e_{v_x}| + \frac{m \,g\,}{k_v}|s_{d\theta}|}{1 - |s_{d\theta}|}\\
    &\doteq e_{v_z}^{stable}
\end{split}
\end{equation}
$|e_{v_z}|$ decreases monotonically to $e_{v_z}^{stable}$. Because $d\theta, e_{v_x}$, asymptotically approach zero, $e_{v_z}^{stable}$ -- and therefore $e_{v_z}$ -- do so as well. We have now shown explicitly that all components of the velocity error $\ve_v$ asymptotically approach zero, so the velocity controller described here asymptotically tracks the desired velocity.

	\bibliographystyle{IEEEtran}
	\bibliography{./references/refs,myRefs} %

}	

\end{document}